\pdfoutput=1

\documentclass[11pt]{article}

\usepackage[final]{acl}

\usepackage{times}
\usepackage{latexsym}

\usepackage[T1]{fontenc}

\usepackage[utf8]{inputenc}

\usepackage{microtype}

\usepackage{inconsolata}

\usepackage{graphicx}
\usepackage{amsmath}
\usepackage{amssymb}
\usepackage{algorithm}
\usepackage{algorithmic}
\usepackage{multirow}
\usepackage{mdframed}
\usepackage{adjustbox}
\usepackage{booktabs}
\usepackage{tcolorbox}
\usepackage{xcolor}
\usepackage{float}
\usepackage{tabularray}
\usepackage{arydshln}
\usepackage{colortbl}
\newmdenv[backgroundcolor=gray!10, linewidth=0pt, innerleftmargin=10pt, innerrightmargin=10pt, innertopmargin=10pt, innerbottommargin=10pt]{highlight}
\definecolor{mypink1}{rgb}{0.858, 0.188, 0.478}

\title{Chain-of-Probe: Examining the Necessity and Accuracy of CoT Step-by-Step}
\author{
  Zezhong Wang$^{1}$\thanks{Work done during internship at Huawei Noah’s Ark Lab.}, Xingshan Zeng$^{2}$\thanks{Corresponding author}, Weiwen Liu$^{2}$, Yufei Wang$^{2}$, Liangyou Li$^{2}$, \\ 
  \bf Yasheng Wang$^{2}$, Lifeng Shang$^{2}$, Xin Jiang$^{2}$, Qun Liu$^{2}$, Kam-Fai Wong$^{1}$ \\
  $^1$The Chinese University of Hong Kong,  $^2$Huawei Noah’s Ark Lab\\
  \texttt{\{zzwang,kfwong\}@se.cuhk.edu.hk}\\
  \texttt{\{zeng.xingshan,liuweiwen8,wang.yufei1,liliangyou\}@huawei.com}\\
  \texttt{\{wangyasheng,Shang.Lifeng,Jiang.Xin,qun.liu\}@huawei.com}
  }

\begin{document}
\maketitle
\begin{abstract}

Current research found the issue of Early Answering in large language models (LLMs), where the models already have an answer before generating the Chain-of-Thought (CoT). This phenomenon suggests a potential lack of necessary dependency between the predicted answer and the reasoning process. Consequently, two important questions arise: (1) Is CoT still necessary if the model already has an answer? (2) Can the correctness of the answer serve as valid evidence for the correctness of CoT?
To address these questions, we propose a method, namely Chain-of-Probe (CoP), to probe changes in confidence during the model's reasoning. The probing results show that in a significant number of question-answer cases, CoT appears to be unnecessary, and this necessity correlates with the simplicity of the task, defined by the reasoning steps required.
Furthermore, by analyzing patterns in confidence change, we examine the correctness of the model's reasoning. Our validation reveals that many responses, although correct in their final answer, contain errors in their reasoning process. To this end, we propose a strategic approach based on CoP to prioritize answers with correct reasoning among multiple candidates, thereby bolstering the reliability of the model's reasoning.
\end{abstract}

\section{Introduction}

Chain-of-Thought (CoT) has been widely proven to effectively improve the accuracy of Large Language Models (LLMs) in reasoning tasks. However, recent research~\cite{lanham2023measuring, turpin2023language} found the issue of Early Answering in LLMs, where LLMs have already predicted an answer before generating the CoT (refer to the left part in Figure 1). This implies that in many cases, the contribution of CoT to the model's final prediction is limited~\cite{lyu-etal-2023-faithful, bentham2024chainofthought, parcalabescu2024measuring, yeo2024interpretable}, or even unnecessary. On the other hand, if the model's predicted answer does not necessarily depend on the CoT, we cannot judge the correctness of the model's reasoning by examining the answers. Even if the model predicts correct answers, it is unreliable if the CoT is incorrect~\cite{zhang2023language, li2024faithful, sui2024fidelis}. 

\begin{figure}[t]
    \centering
    \includegraphics[width=\linewidth]{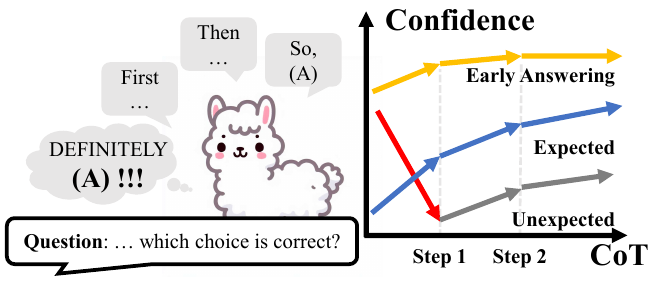}
   \vspace{-0.3cm}
    \caption{Diagram of early answering and Chain-of-Probe. Line graphs illustrate several typical patterns of confidence changes.}
    \label{fig:intro}
\vspace{-0.3cm}
\end{figure}

In order to effectively evaluate the necessity and accuracy of CoT, we propose a probing method, namely Chain-of-Probe (CoP), to detect the changes in the model's thought during reasoning. Specifically, after the model completes each step of reasoning, we ask it to output a prediction based on the current reasoning and record its confidence. 
We believe that the changes in confidence can help us understand the impact of each step of reasoning on the model's decision-making process~\cite{wang2024chainofthoughtreasoningprompting}. As shown in the right side of Figure~\ref{fig:intro}, these confidence trends across three different reasoning processes illustrate this phenomenon.

Based on CoP, we investigate the issue of early answering, intending to analyze the underlying causes of early answering, thereby determining when the CoT is necessary. 
We first conduct statistical analysis on multiple-choice reasoning datasets.
Results show that all LLMs exhibit the issue of early answering (i.e., choosing the same answer throughout the entire reasoning process) on a large number of question-answer cases. 
More surprisingly, the accuracy of the predictions when the models exhibit early answering is significantly higher than the accuracy obtained when changing their choices during the reasoning process (generally exceeding 20\%). 
This seems to contradict the view that CoT can improve model performance.

We further analyze the correlation between early answering and accuracy and find that early answering is linked to question difficulty: the model tends to predict answers in advance for simpler questions~\cite{madaan-etal-2023-makes}. This suggests that CoT is often unnecessary for simple questions. For challenging questions, CoT is more likely to alter the model's initial choice, though not always positively. Our analysis also indicates that higher model confidence during reasoning correlates with a higher likelihood of correctness. Based on this, we propose the CoP Score to evaluate and select CoTs, aiming for more positive improvements. Experiments demonstrate that selecting responses based on the CoP Score achieves accuracy comparable to majority voting.

Selection based on the CoP score is relatively optimal among the candidates. However, it does not guarantee the correctness of the processes within CoT~\cite{wang-etal-2023-towards}. The correctness of CoT is equally crucial as part of the model's output, reflecting the reliability of the answers~\cite{bao2024llmschainofthoughtnoncausalreasoners, ye2022unreliabilityexplanationsfewshotprompting, jung-etal-2022-maieutic}. Our further experiments reveal that about 20\% of CoTs with correct answers include reasoning errors. In subsequent case studies combining CoP, we observe a significant decrease in confidence when the model makes incorrect steps in its reasoning process. Based on this insight, we extract features from CoP to train a decision tree, called CoP Tree, which is used to identify CoT that may contain errors.
We propose to resample if errors are identified by the CoP Tree.
Such an inference method improves the model's overall accuracy by 13\% on average across different models when evaluating and also considering CoT correctness.

We summarize our contribution into the following three key points: 
\vspace{-0.2cm}
\begin{itemize}
    \item We propose a novel method, Chain-of-Probe (CoP), to detect changes in model confidence.\vspace{-0.2cm}
    \item We identify that the problem of early answering in the model is due to the simplicity of the questions, making CoT unnecessary. \vspace{-0.2cm}
    \item We find that the changing pattern of confidence during the model's reasoning can be used to examine the correctness of the model's CoT and answers, thus improving overall accuracy.\vspace{-0.2cm}
\end{itemize}

\section{Related Works}

\begin{figure*}[t]
    \centering
    \vspace{-0.3cm}
    \includegraphics[width=\linewidth]{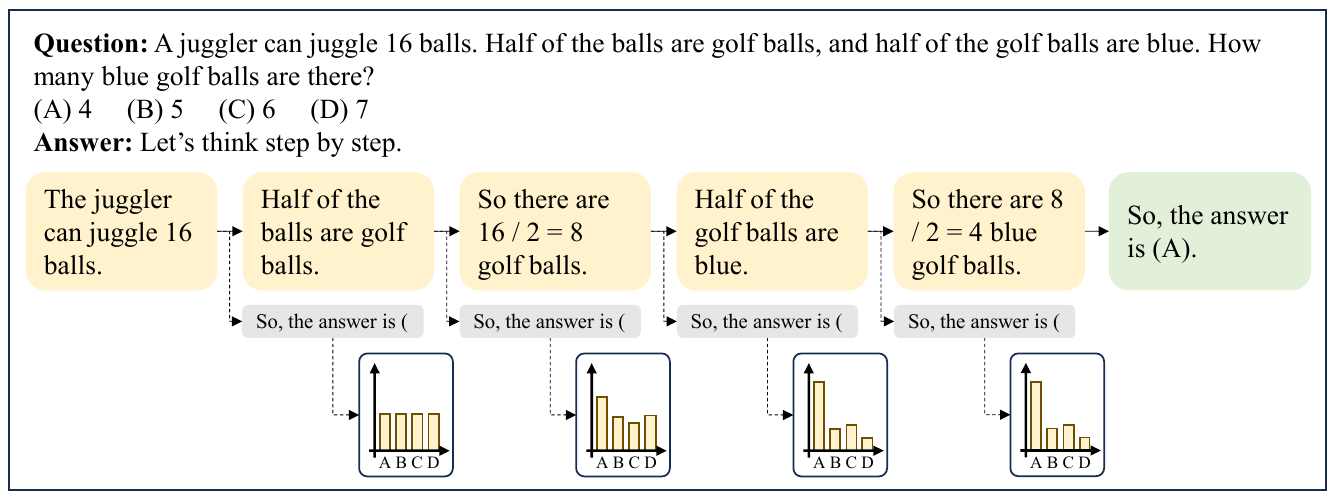}
    \vspace{-0.3cm}
    \caption{The pipeline of Chain-of-Probe with a running example. Yellow boxes represent each reasoning step in the CoT. Gray boxes denote predefined probing strings. In this case, \{A, B, C, D\} serves as the target token set, with each probing collecting the model's predicted probabilities for these four tokens (illustrated by yellow bar charts).}
    \label{fig:pipeline}
    \vspace{-0.3cm}
\end{figure*}

Extensive research has shown that CoT techniques significantly improve the reasoning abilities of LLMs~\cite{wei2023chainofthought}. However, \citet{lanham2023measuring} and \citet{turpin2023language} highlights an issue known as early answering, where models arrive at the answer before the CoT is fully generated. Their experiments found that even if the CoT is forcibly interrupted or erroneous information is added, the model still arrives at the same answer. 
Similarly, some work~\cite{parcalabescu2024measuring, paul2024making} found that LLMs are not always faithful in using their intermediate reasoning steps when generating answers. \citet{li2024faithful} discovered that while the CoT sometimes misses critical context, the model often recalls this information directly from the context when answering. These observations suggest that LLMs often answer questions without depending on CoT \cite{lyu-etal-2023-faithful, bentham2024chainofthought, yeo2024interpretable, sprague2024cotcotchainofthoughthelps}.

Besides, as the model's final prediction might not necessarily depend on the generated CoT, evaluating the accuracy of the CoT based on the final answer becomes inaccurate. The quality of the CoT, as part of the model's response, is equally important~\cite{lightman2023lets, jacovi2024chainofthoughtstrongweakestlink}. Deriving correct answers from flawed reasoning remains unreliable~\cite{zhang2023language, bao2024llmschainofthoughtnoncausalreasoners}. As a result, increasing efforts focus on improving the accuracy of the model's reasoning process rather than just the accuracy of the prediction results~\cite{radhakrishnan2023question}. FiDeLiS~\cite{sui2024fidelis} uses an external knowledge graph to enhance CoT accuracy. Another, the Selection-Inference (SI) model \cite{creswell2022faithful}, includes a value function to assess the quality of reasoning steps, guiding beam search to improve reasoning quality. Additionally, there are some methods~\cite{ji-etal-2023-towards, zheng2023progressivehint} that allow models to recheck their reasoning and correct any errors after generating CoT to ensure its accuracy. While these methods improve CoT accuracy, they introduce significant additional costs, such as requiring an external module or an extra round of inference.

\section{Methodology}


In this section, we propose a Probing method, namely, CoP. Essentially, after the model generates each step of reasoning, we prompt it to predict an answer and record the corresponding confidence. Figure~\ref{fig:pipeline} illustrates the process of CoP.

We denote the input to the model as $x$, i.e., the prompt, 
and the $i$-th step in the CoT generated by the model as $s_i$. 
We also introduce a probing string $a_*$, 
such as, \textit{" So, the answer is ("}. We expect the model to predict an answer after concatenating $a_*$ to each step, like a choice index \textit{A} or \textit{B} for multiple-choice questions. 
We then define a target token set $\mathcal{V}$ that includes all the possible answers, such as \{A, B, C, D\}, etc. This set needs to be customized based on the datasets that are specifically used.
After the model generates the $i$-th step of reasoning $s_i$, we concatenate $a_*$ to $s_i$, using the model to predict the word distribution of the next token:
\begin{equation}
\label{eq:llm_softmax}
    P(y|x; s_{1:i} ;a_*) = \mathrm{softmax}\ \mathrm{LLM}(x; s_{1:i}; a_*)
\end{equation}
We extract the probabilities corresponding to the tokens defined in the set $\mathcal{V}$ from the word distribution, which serves as the confidence set for the $i$-th step encompassing all possible final predictions:
\begin{equation}
\label{eq:confidence_set}
\begin{aligned}
c_i &= \left[ p_i^v \mid v \in \mathcal{V}\right] \\
\text{where } p_i^v &= P(y = v \mid x; s_{1:i}; a_*)
\end{aligned}
\end{equation}
Notably, we also do a probe before the model's reasoning (i.e., the model only read the prompt). The result of this initial probe is denoted as $c_0$, which can be regarded as the direct answer without any CoT. Then, we divide the reasoning process according to sentences, and after each sentence, we probe once based on Equation (\ref{eq:llm_softmax}) and (\ref{eq:confidence_set}). At the end of the CoT generation, we can collect a confidence matrix $c=[c_0, \cdots,c_k]\in\mathbb{R}^{(k+1)\times|\mathcal{V}|}$, where $k$ is the number of reasoning steps.


In practice, we propose a \textbf{cache fallback} algorithm to reduce redundant calculations, thereby conserving substantial computational resources.
Specifically, we only maintain the Key-Value (KV) cache \cite{vaswani2017attention} generated during the CoT process, denoted as $\mathcal{M}$. The KV cache generated after the $i$-th reasoning step is $\mathcal{M}_i = \{m_x, m_{s_1}, \cdots, m_{s_i}\}$, where $m_{(*)}$ represents the cache produced by the corresponding sequence encoding. During probing, $m_{a_*}$ is generated and temporarily added to KV cache, $\mathcal{M}_i \leftarrow \mathcal{M}_i \cup \{m_{a_*}\}$. After obtaining the probing result, $m_{a_*}$ is removed from $\mathcal{M}_i$, allowing the process to revert from probing back to the CoT generation and continue with the next step. In this way, we circumvent redundant calculations, thereby conserving substantial computational resources. The additional computational overhead during the probing process is reduced to the generation of $k \times l$ tokens, where $l$ represents the length of the probing string.


\section{Experiment}
\label{sec:experiments}

\begin{figure*}[t]
    \centering
    \vspace{-0.2cm}
    \includegraphics[width=0.95\linewidth]{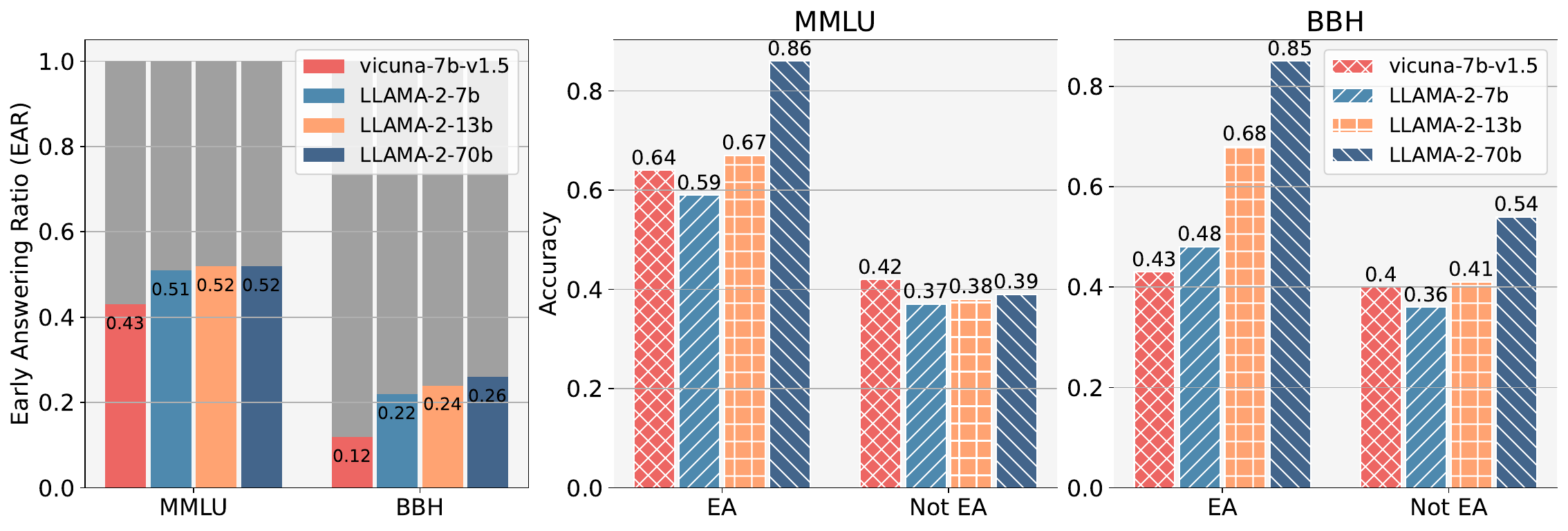}
    \vspace{-0.3cm}
    \caption{The left figure shows the early answering ratio in the model on the MMLU and BBH datasets. The right two figures compare the model's accuracy on these datasets, distinguishing between cases with early answering (EA) issues and those without (Not EA).}
    \label{fig:ear_acc}
\vspace{-0.3cm}
\end{figure*}

\subsection{Settings}

We mainly conducted experiments on three reasoning datasets: MMLU~\cite{hendrycks2021measuring}, BBH~\cite{suzgun2022challenging}, and ARC~\cite{clark2018think}. For specific dataset information and reasoning settings, please refer to Appendix~\ref{sec:settings_app}.


The models used in the experiments include LLaMA-2 (7B, 13B, 70B)~\cite{touvron2023llama}, Vicuna 7B~\cite{vicuna2023}, Mistral 7B~\cite{jiang2023mistral}, LLaMA-3 8B~\cite{meta_llama_3}, and Qwen2 7B~\cite{qwen2} (Refer to Appendix~\ref{sec:checkpoints} for detailed checkpoint versions). 

In the following sections, 
we use CoP to first analyze the conditions in which the LLM is more prone to experiencing the early answering issue.
Next, we examine the correlation between model confidence and accuracy, and design a candidate answer selection strategy, validating it across three datasets. Finally, through case analysis and pattern recognition, we identify abnormal changes in model confidence during reasoning, allowing us to exclude CoT with potential reasoning errors.

\subsection{Early Answering Issue Analysis}

In this experiment, we systematically conducted a statistical analysis of the early answering issue in open-source models and examined the causes.

\subsubsection{Early Answering Criterion}

We define the early answering issue as the model consistently choosing the same answer throughout the entire reasoning process. 
Given the confidence matrix $c=[c_0, \cdots, c_k]$ obtained from probing, we take the index of the highest probability in each row as the prediction for that step, i.e.,
\begin{equation}
\begin{aligned}
    \hat{j} &= [j_0, \ldots, j_k], \\
    \textrm{where } j_i &= \arg\max_{j} (c_{ij} \mid j = 1, \ldots, |\mathcal{V}|).
\end{aligned}
\label{eq:ea_c1}
\end{equation}
If the prediction for each step matches the model's final prediction, we consider that the model has not changed its choice during the reasoning, indicating an early answering issue. The process can be described as follows: 
\begin{equation}
    y = 
\begin{cases} 
1 & \text{if } j_i = j^* \text{ for all } i \in \{0, \ldots, k\} \\
0 & \text{otherwise}
\end{cases}
\label{eq:ea_c2}
\end{equation}
where $y=1$ indicates that the early answering issue occurred, and $y=0$ indicates that it did not.

\begin{figure*}[t]
    \centering
    \includegraphics[width=0.98\linewidth]{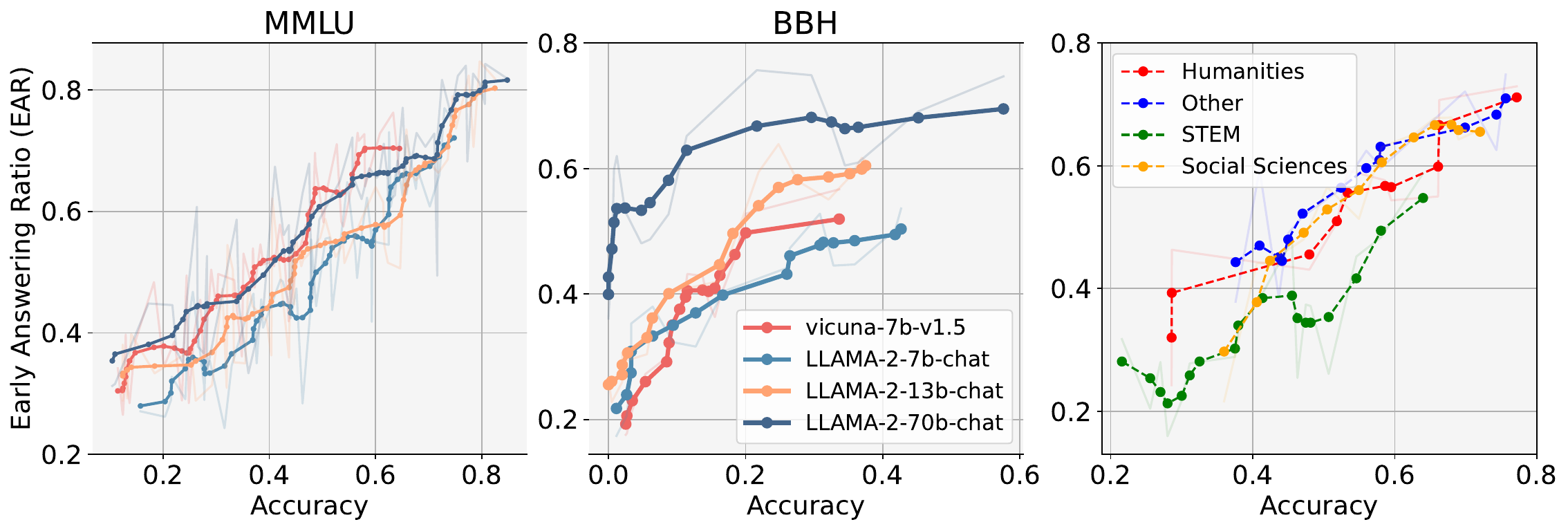}
    \vspace{-0.1cm}
    \caption{Relationship between EAR and accuracy. The two graphs on the left show the results of four models on the MMLU and BBH datasets. The right part shows the results of the LLaMA-2 7b model on the MMLU dataset, categorized by disciplines. Gaussian smoothing, with sigma=1 and order=0, was applied to each line to better observe overall trends.}
    \label{fig:domain_ear_acc}
\vspace{-0.3cm}
\end{figure*}

\subsubsection{Early Answering Statistical Analysis}

Based on this criterion, we conduct experiments on four models across three datasets and reported the \textbf{E}arly \textbf{A}nswering \textbf{R}atio (\textbf{EAR}), indicating the percentage of instances where the model exhibits early answering issues.

From the first sub-figure of Figure~\ref{fig:ear_acc}, we find that early answering is a common issue across all models instead of an isolated incident. For example, on MMLU, multiple models exhibit early answering in nearly half of the cases, indicating that half of the CoTs might be unnecessary.
We further analyze the accuracy of the models both when early answering occurs and when it does not, as shown in Figure~\ref{fig:ear_acc}. Surprisingly, 
when the model changes its decision after generating a CoT, its accuracy tends to be lower.
This contradicts the belief that CoT enhances performance.

\subsubsection{Cause of Early Answering}

To investigate the reasons behind the \textit{"negative effect of CoT,"}, we further analyze the relationship between EAR and accuracy and plot the experimental results of the MMLU and BBH datasets on Figure~\ref{fig:domain_ear_acc}. 
From the left two sub-figures, we can observe a positive correlation between EAR and accuracy across all models in both datasets. This observation is consistent with our previous finding that models achieve higher accuracy when the early answering issue occurs. If we use accuracy as an indicator of task difficulty — i.e., higher accuracy indicating that the task is relatively easy for the model, and vice versa — we can draw the following finding:

\begin{highlight}
\textit{\textbf{Finding 1:} For reasoning-based tasks, easier questions (i.e., those with higher accuracy) are more likely to lead to early answering. This suggests that CoT may not be necessary for those easy tasks.}
\end{highlight}

Intuitively, simpler questions require less reasoning steps and can often be answered directly.
We further categorize the results of LLaMA-2 7b on the MMLU dataset to investigate the relationship between EAR and Accuracy across various disciplines. The results are shown in the right sub-figure of Figure~\ref{fig:domain_ear_acc}. 
Overall, the trend observed across disciplines remains consistent: higher accuracy correlates with a higher EAR. By comparing the curves of different disciplines, we find that STEM, in particular, shows a lower EAR compared to others (with an average EAR of 34.6\% for STEM tasks and 51\% overall), which implies that STEM may require more reasoning compared to other disciplines.
This observation also aligns with our intuition. 
We conducted similar experiments on datasets with human-annotated difficulty levels and reached similar conclusions, with details provided in Appendix~\ref{sec:app_arc_ear}. Additionally, we present the specific numerical values for the correlation analysis between EAR and Accuracy in Appendix~\ref{sec:corr_ana_app}.


\begin{figure}[t]
    \centering
    \vspace{-0.3cm}
    \includegraphics[width=\linewidth]{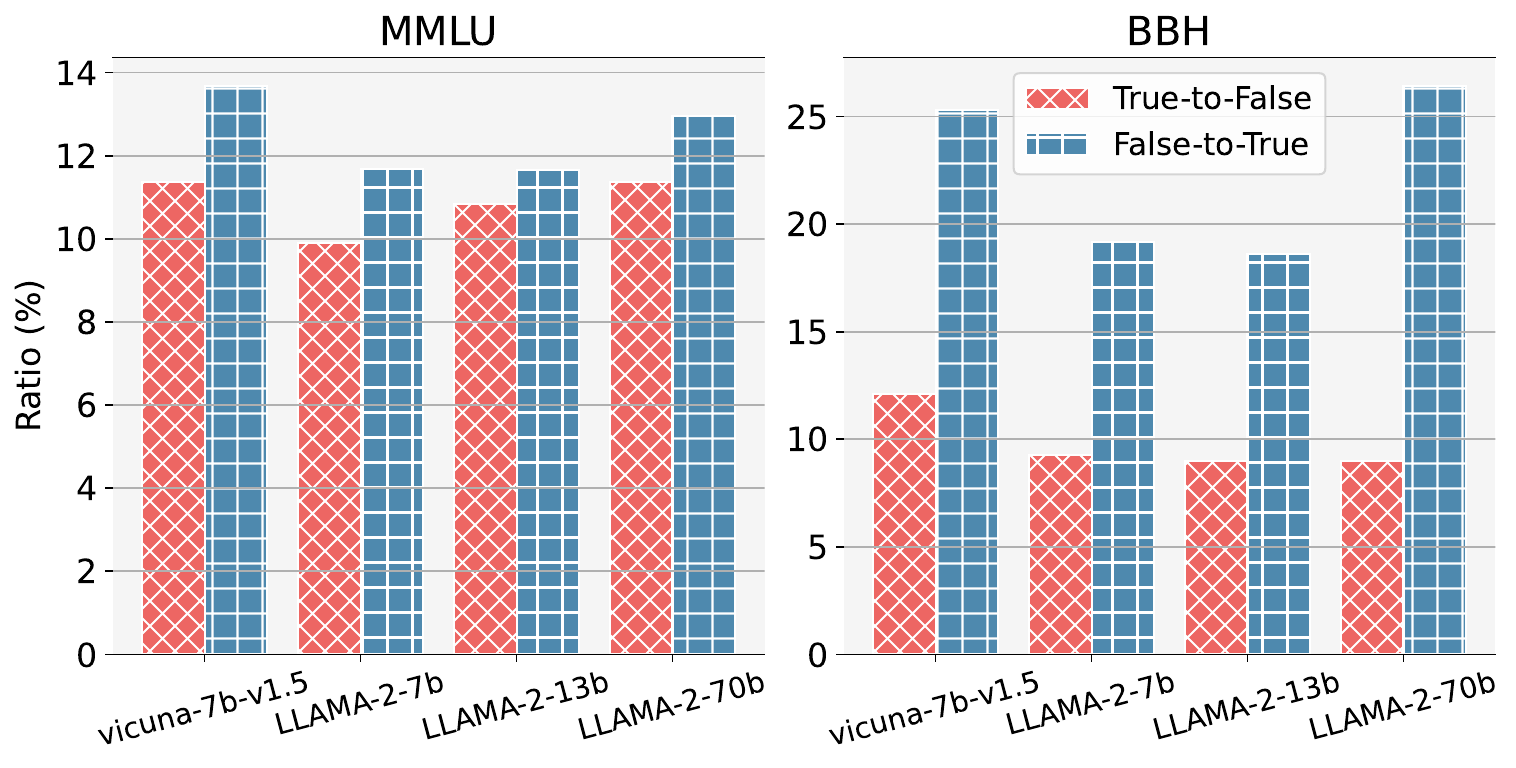}
    \vspace{-0.3cm}
    \caption{The ratio of CoT changing answers from True to False versus from False to True. }
    \label{fig:T2FandF2T}
\vspace{-0.5cm}
\end{figure}

In essence, the questions where CoT can be effective tend to be more challenging, resulting in naturally lower accuracy, rather than CoT having negative effects.
In fact, when CoT influences model decisions, there are generally more correct adjustments than incorrect ones on a macro scale (refer to Figure~\ref{fig:T2FandF2T} and Appendix~\ref{sec:pos_eng_cot}). This is why CoT can enhance model performance. 
It is undeniable that there are cases where the model makes mistakes after reasoning, which suggests that CoT may not always be beneficial~\cite{singhal2022largelanguagemodelsencode, gupta2024whispersdoubtamidstechoes, wang2024mmluprorobustchallengingmultitask}.
Hence, could we discern whether a CoT is a positive or negative process, thus eliminating CoTs that lead to incorrect adjustments? We explore the answer in the next two subsections.

\begin{table*}[t]
\centering
\begin{adjustbox}{max width=\textwidth}
\begin{tabular}{llllllllc}
\hline
\multicolumn{1}{c}{\textbf{Acc (\%)}} & \multicolumn{1}{c}{\textbf{Vicuna 7b}} & \multicolumn{1}{c}{\textbf{Mistral 7b}} & \multicolumn{1}{c}{\textbf{LLaMA-3 8b}} & \multicolumn{1}{c}{\textbf{Qwen2 7b}} & \multicolumn{1}{c}{\textbf{LLaMA-2 7b}} & \multicolumn{1}{c}{\textbf{LLaMA-2 13b}} & \multicolumn{1}{c|}{\textbf{LLaMA-2 70b}} & \textbf{Avg.}             \\ \hline
\rowcolor[HTML]{EFEFEF}\multicolumn{9}{c}{\textbf{\textit{MMLU}}}                                                                                                                                                                                                                                                                                                                                      \\
\textbf{GS}                           & 52.5                                   & 62.7                                    & 67.2                                    & 70.5                                  & 49.7                                    & 55.9                                     & \multicolumn{1}{l|}{69.0}                 & 61.1                     \\
\textbf{Maj@5}                        & 53.6                                   & \textbf{64.5}                           & 69.1                                    & \textbf{73.0}                         & 51.3                                    & 57.7                                     & \multicolumn{1}{l|}{\textbf{70.6}}        & 62.8                     \\
\textbf{CoPS@5}                          & \textbf{54.7 \textcolor{mypink1}{(+1.1)}}                   & 64.4 (-0.1)                             & \textbf{70.2 \textcolor{mypink1}{(+1.1)}}                    & 72.1 (-0.9)                           & \textbf{52.7 \textcolor{mypink1}{(+1.4)}}                    & \textbf{59.0 \textcolor{mypink1}{(+1.3)}}                     & \multicolumn{1}{l|}{69.9 (-0.7)}                 & \textbf{63.3}            \\
\rowcolor[HTML]{EFEFEF}\multicolumn{9}{c}{\textbf{\textit{BBH}}}                                                                                                                                                                                                                                                                                                                                       \\
\textbf{GS}                           & 45.7                                   & 56.5                                    & 71.4                                    & 65.7                                  & 44.1                                    & 48.2                                     & \multicolumn{1}{l|}{63.1}                 & 56.4                     \\
\textbf{Maj@5}                        & 46.9                                   & \textbf{59.2}                           & \textbf{74.0}                           & 68.7                                  & 44.6                                    & 49.6                                     & \multicolumn{1}{l|}{68.5}                 & 58.8                     \\
\textbf{CoPS@5}                          & \textbf{48.5 \textcolor{mypink1}{(+1.6)}}                   & 58.9 (-0.3)                             & 73.4 (-0.6)                             & \textbf{70.1 \textcolor{mypink1}{(+1.4)}}                  & \textbf{45.5 \textcolor{mypink1}{(+0.9)}}                    & \textbf{50.8 \textcolor{mypink1}{(+1.2)}}                     & \multicolumn{1}{l|}{\textbf{69.5} \textcolor{mypink1}{(+1.0)}}        & \textbf{59.5}            \\
\rowcolor[HTML]{EFEFEF}\multicolumn{9}{c}{\textbf{\textit{ARC-Easy}}}                                                                                                                                                                                                                                                                                                                                     \\
\textbf{GS}                           & 81.0                                   & 89.4                                    & 93.5                                    & 94.9                                  & 83.0                                    & 88.4                                     & \multicolumn{1}{l|}{93.4}                 & 89.1                     \\
\textbf{Maj@5}                        & 83.3                                   & 90.9                                    & 94.2                                    & 95.7                                  & 84.5                                    & 88.7                                     & \multicolumn{1}{l|}{94.5}                 & 90.3                     \\
\textbf{CoPS@5}                          & \textbf{84.3 \textcolor{mypink1}{(+1.0)}}                   & \textbf{92.2 \textcolor{mypink1}{(+1.3)}}                    & \textbf{95.2 \textcolor{mypink1}{(+1.0)}}                    & \textbf{96.9 \textcolor{mypink1}{(+1.2)}}                  & \textbf{84.8 \textcolor{mypink1}{(+0.3)}}                    & \textbf{89.0 \textcolor{mypink1}{(+0.3)}}                     & \multicolumn{1}{l|}{\textbf{95.8} \textcolor{mypink1}{(+1.3)}}        & \textbf{91.2}            \\
\rowcolor[HTML]{EFEFEF}\multicolumn{9}{c}{\textbf{\textit{ARC-Challenge}}}                                                                                                                                                                                                                                                                                                                                     \\
\textbf{GS}                           & 66.9                                   & 82.2                                    & 86.1                                    & 88.7                                  & 69.2                                    & 73.2                                     & \multicolumn{1}{l|}{84.0}                 & 78.6                     \\
\textbf{Maj@5}                        & 72.2                                   & 82.8                                    & 87.1                                    & 91.0                                  & 71.4                                    & 77.4                                     & \multicolumn{1}{l|}{88.1}                 & 81.4                     \\
\textbf{CoPS@5}                          & \textbf{73.0 \textcolor{mypink1}{(+0.8)}}                   & \textbf{84.1 \textcolor{mypink1}{(+1.3)}}                    & \textbf{87.4 \textcolor{mypink1}{(+0.3)}}                    & \textbf{91.3 \textcolor{mypink1}{(+0.3)}}                  & \textbf{73.3 \textcolor{mypink1}{(+1.9)}}                    & \textbf{78.4 \textcolor{mypink1}{(+1.0)}}                     & \multicolumn{1}{l|}{\textbf{88.4} \textcolor{mypink1}{(+0.3)}}        & \textbf{82.3}            \\ 
\cdashline{1-9} \textbf{Avg.}                                  & +1.1                                   & +0.6                                    & +0.5                                    & +0.5                                  & +1.1                                    & +0.9                                     & +0.5                                      & \multicolumn{1}{l}{+0.7} \\ 
\hline
\end{tabular}
\end{adjustbox}
\caption{The comparison results of CoT selection. GS, Maj@5, and $S$@5 indicate the settings of greedy search, 5-time sampling with majority vote, and 5-time sampling with the CoP score (our method), respectively. The best result is highlighted in \textbf{bold}.The improvements of $S$@5 compared with Maj@5 are listed in parentheses, with the positive improvements marked in \textcolor{mypink1}{red}. The last row shows the average improvement of $S$@5 compared to Maj@5.} 
\label{tab:cot_score}
\vspace{-0.3cm}
\end{table*}

\subsection{Positive CoTs discrimination}

\subsubsection{CoP Score}

To select positive CoT, we attempt to identify indicators from probing results that evaluate the quality of CoT and assign corresponding scores. According to Equation (\ref{eq:ea_c1}) and (\ref{eq:ea_c2}), early answering essentially refers to the model maintaining high confidence in its predictions throughout the entire reasoning process. We hypothesize that there is also a correlation between confidence and accuracy, such as higher confidence in reasoning being more likely to yield accurate answers. To verify this, we track the confidence associated with the final prediction $v^*$ at each step of probing to observe the trend of the model's confidence during reasoning. Then, we defined a metric, namely \textbf{CoP Score} as:
\begin{equation}
\mathrm{CoPS} = \frac{1}{k+1}\sum_{i=0}^{k} p_i^{v^*} + \frac{1}{k}\underbrace{(p_k^{v^*} - p_0^{v^*})}_{\sum_{i=1}^{k} \left(p_i^{v^*} - p_{i-1}^{v^*}\right)}
\end{equation}
The first term measures the model's average confidence during the reasoning process, while the second evaluates the average change in confidence. Intuitively, higher confidence or greater increases in confidence suggest that CoT benefits the final prediction, making these CoTs more likely to have a positive impact.

To further verify whether the CoP Score accurately reflects the quality of CoT, we perform a correlation analysis between the CoP Score and Accuracy.
Based on the probing results, we compute a CoP Score for each CoT and sorted them accordingly. Then, we partition the sorted CoT set into 10 sections and calculate the accuracy within each section. We plot a line graph in Figure~\ref{fig:cot_score_acc} using the accuracy and average CoP Score of each section as coordinates.

Figure~\ref{fig:cot_score_acc} reveals a clear positive correlation between the CoP score and accuracy (Refer to Appendix~\ref{sec:corr_ana_app} for details). As the CoP score essentially represents model confidence, we can further draw the following finding:

\begin{figure}[t]
    \centering
    \vspace{-0.3cm}
    \includegraphics[width=\linewidth]{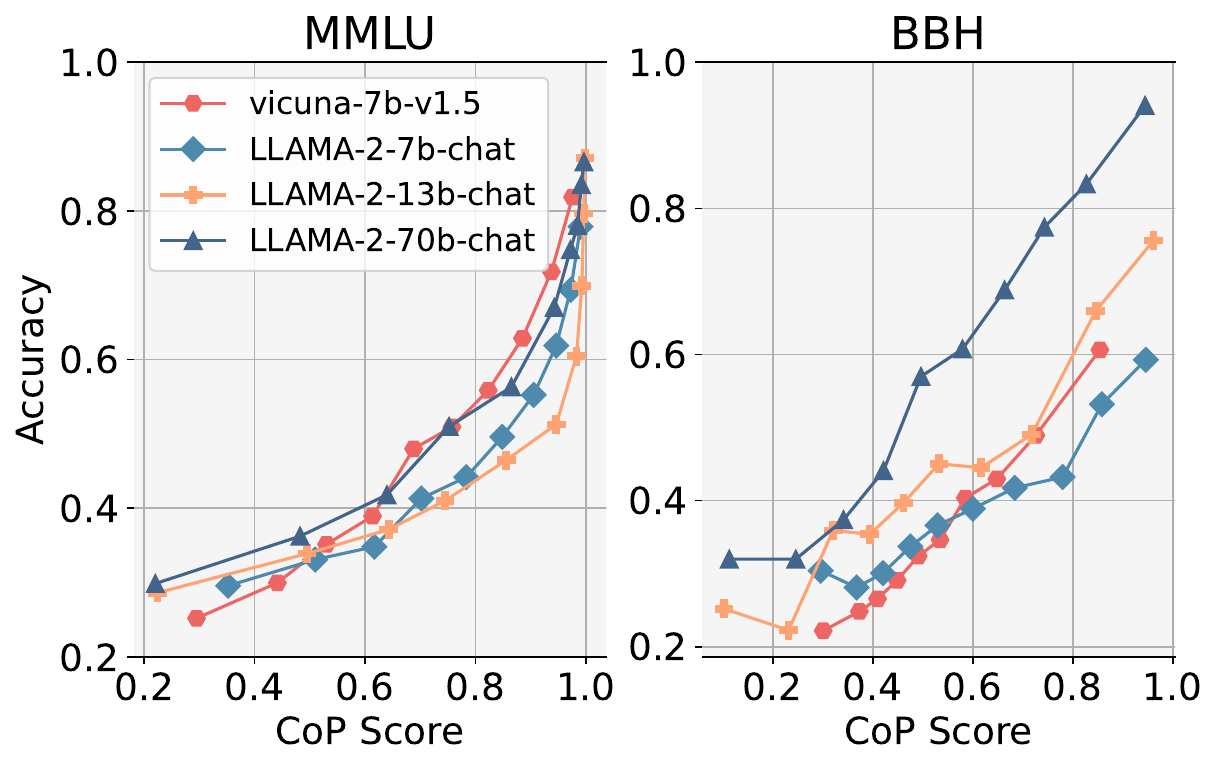}
    \vspace{-0.3cm}
    \caption{The correlation between CoP Score and Accuracy. Positive correlation is found between them.}
    \label{fig:cot_score_acc}
\vspace{-0.5cm}
\end{figure}

\begin{highlight}
\textit{\textbf{Finding 2:} The model is confident during the inference process (high CoP Score), often resulting in accurate predictions.}
\end{highlight}

This result also validates our hypothesis: using the CoP Score to guide the selection of better CoTs is potentially feasible.

\subsubsection{CoT Selection Strategy}

Based on the correlation analysis, we evaluate CoP Score's ability to select CoT across three datasets: MMLU, BBH, and ARC. For each question, we have LLMs sample five responses and extract the corresponding CoP.
Then, we calculate the CoP Score for each response based on the CoP and select the response with the highest CoP Score as the final prediction. We compare our method ($S$@5) against five-time sampling with majority voting (Maj@5) and greedy search (GS). 
The results in Table~\ref{tab:cot_score} show that choosing answers based on the CoP score yields marginally better accuracy compared to using multiple voting. Compared to Maj@5, S@5 achieves an average improvement of over 1\% across several models (Vicuna 7b and LLaMA-2 7b). 
We further conducted a one-tailed paired t-test to compare CoPS with the baseline method on the ARC-Challenge dataset. Our method achieved consistently higher accuracy (Mean=82.27\%) compared to the baseline (Mean=81.43\%). The analysis revealed that this improvement was statistically significant (p < 0.005) with a large effect size (Cohen's d = 1.38), indicating that our method provides a reliable and meaningful enhancement over the baseline approach.
When compared across different models, selecting answers based on the CoP Score yields even greater improvements for weaker models (earlier or smaller models).

Notably, selection based on the CoP score is relatively optimal among the candidates. While it is more likely to contain the correct answer and the model has higher confidence in it, we still cannot ensure that every reasoning step is accurate. As part of the model output, the correctness of the CoT is equally important. Obtaining numerous correct answers based on flawed reasoning remains unreliable. 
To delve deeper into the correctness of each step within CoT, 
we transition from macro-level correlation analysis to examining the underlying patterns of confidence.
\begin{figure}[t]
    \centering
    \includegraphics[width=\linewidth]{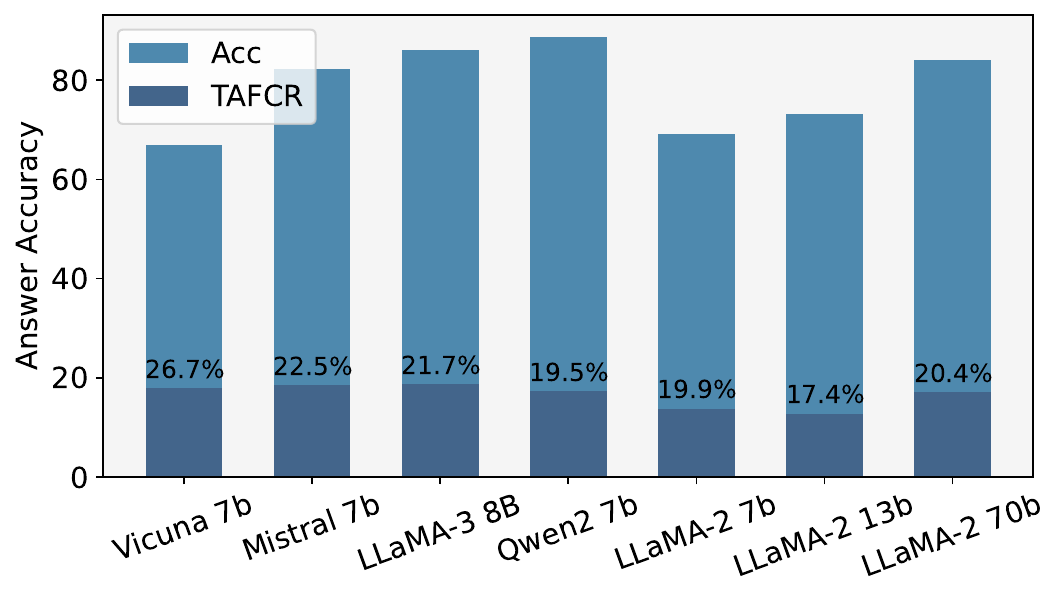}
    \caption{Answer Accuracy and True Answer False CoT Ratio (TAFCR).}
    \label{fig:tafcr_bar}
\vspace{-0.5cm}
\end{figure}


\begin{figure*}[t]
    \centering
    \includegraphics[width=\linewidth]{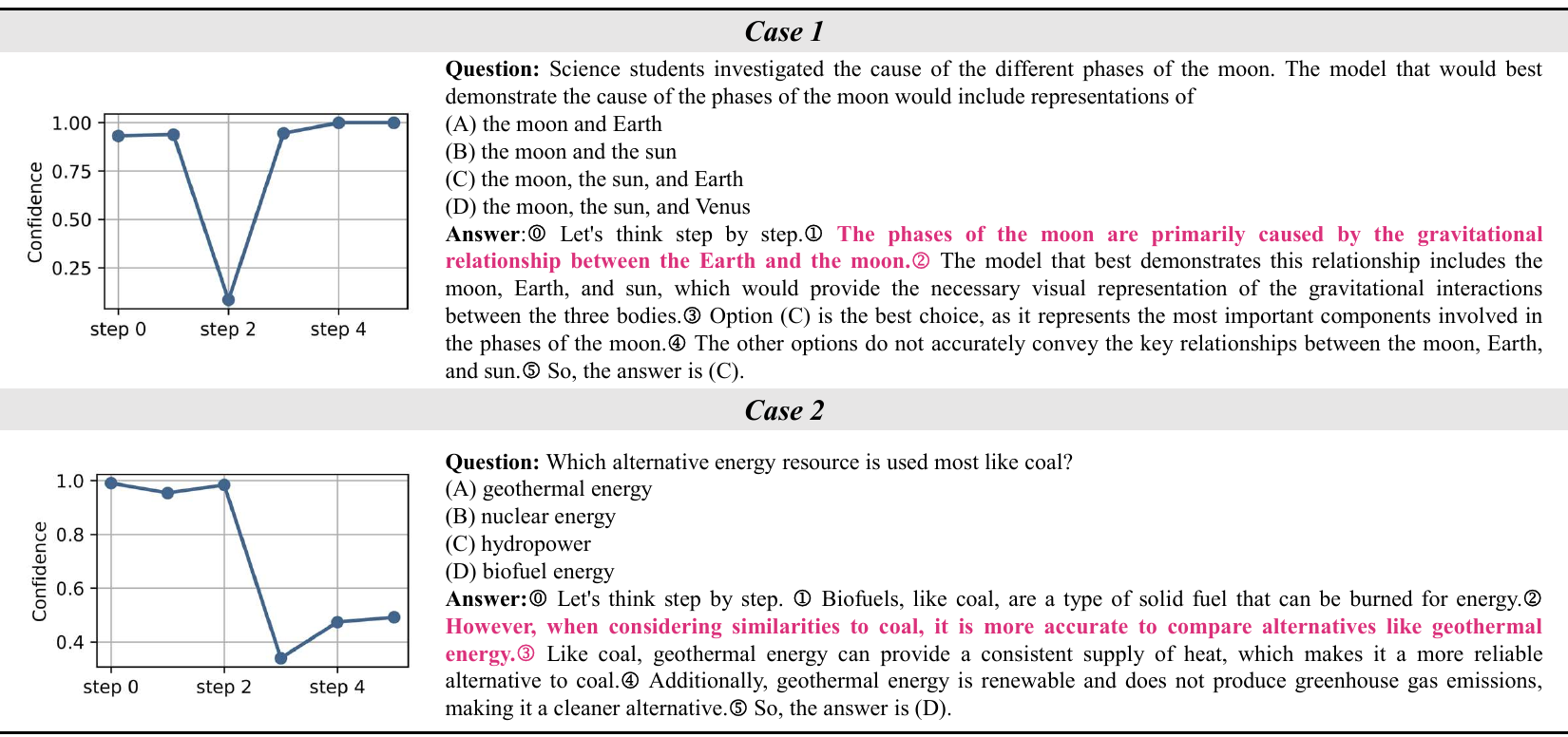}
    \caption{Analysis of two real cases: The line plot shows changes in the model's confidence in its final prediction over the reasoning steps. On the right are the corresponding cases, including the question and the response generated by the model. We add numerical indices to indicate probing points, helping readers match the line plot with the CoT. The step causing the significant drop is marked in \textcolor{mypink1}{red}.}
    \label{fig:case_analysis}
\end{figure*}

\subsection{Reasoning Step Correctness Verification}
\subsubsection{True Answer, False CoT}

We first investigate the ratio of responses with correct answers but false CoT. We collect responses generated by the seven models for the questions in the ARC-Challenge dataset. Then we ask GPT-4~\cite{openai2024gpt4} to check whether those responses are correct, considering any CoT with even a single error as incorrect (refer to Appendix~\ref{sec:gpt4-eval} for details). To ensure the reliability of GPT-4's evaluation, we randomly selected 50 samples from the responses of all models on ARC-Challenge for manual annotation. We recruited four PhD students to annotate the golden answers of whether both CoT and final answers are correct. All the answers are consistently validated by all four annotators.
The Pearson correlation coefficient between GPT-4's annotations and the golden answers is \textbf{0.879}, proving that GPT-4 is a capable evaluator. We define a metric, namely \textbf{T}rue \textbf{A}nswer with \textbf{F}alse \textbf{C}oT \textbf{R}atio (\textbf{TAFCR}), to evaluate this situation:
\begin{equation}
    \mathrm{TAFCR} = \frac{\#\mathrm{TA\cap FC}}{\#\mathrm{TA}}
\end{equation}
We report TAFCR in Figure~\ref{fig:tafcr_bar}. The results show that around 20\% of the "correct" responses contain errors in their CoT. However, according to the Exact Match (EM) metric for answers, these cases are marked as correct in the current benchmark. We summarize the following findings:

\begin{table*}[t]
\centering
\renewcommand{\arraystretch}{1.1}
\begin{adjustbox}{max width=\textwidth}
\begin{tabular}{lcccccccc}
\hline
\textbf{Metrics}        & \textbf{Vicuna 7b} & \textbf{Mistral 7b} & \textbf{LLaMA-3 8B} & \textbf{Qwen2 7b} & \textbf{LLaMA-2 7b} & \textbf{LLaMA-2 13b} & \textbf{LLaMA-2 70b} & \textbf{Avg.} \\ \hline
\rowcolor[HTML]{EFEFEF}\multicolumn{9}{c}{\textit{CoT Accuracy (\%)}}                                                                                                                                                        \\
\textbf{GS}        & 49.1               & 63.7                & 67.4                & 71.4              & 55.4                & 60.5                 & 66.9                 & 62.0          \\
\textbf{CoPS@3}    & 55.3               & 70.7                & 70.6                & 75.3              & 60.1                & 65.3                 & 76.8                 & 67.7          \\
\textbf{L2M}       & 50.0               & 71.9                & 70.7                & 78.7              & 59.1                & 68.5                 & 74.5                 & 67.6          \\
\textbf{ToT}       & 48.7               & \textbf{78.6}       & \textbf{79.2}       & 81.2              & 57.3                & 66.7                 & 79.5                 & 70.2          \\
\textbf{CoPT}      & \textbf{67.6}      & 76.4                & 78.9                & \textbf{82.8}     & \textbf{68.7}       & \textbf{72.8}        & \textbf{81.1}        & 75.5          \\
\rowcolor[HTML]{EFEFEF}\multicolumn{9}{c}{\textit{CoPT Classification Performance (\%)}}                                                                                                                                     \\
\textbf{Precision} & 84.1               & 89.6                & 91.0                & 91.2              & 81.4                & 87.5                 & 91.1                 & 88.0          \\
\textbf{Recall}    & 67.2               & 69.3                & 70.3                & 70.6              & 72.5                & 67.1                 & 71.1                 & 69.7          \\
\textbf{F1}        & 74.7               & 78.1                & 79.3                & 79.6              & 76.7                & 76.0                 & 79.8                 & 77.7          \\
\cdashline{1-9}\textbf{\# Samples}     & 4.17               & 3.89                & 3.53                & 3.12              & 4.34                & 3.87                 & 4.03                 & 3.89          \\ \hline
\end{tabular}
\end{adjustbox}
\caption{Results of reasoning step correctness verification on ARC-Challenge. TA-FCR stands for True Answer with False CoT Ratio. GS, CoPS@3, L2M, ToT, and CoPT represent the CoT accuracy under the greedy search, 3-time sampling with the CoP score, Least-to-Most Prompting, Tree-of-Thought, and CoP Tree resampling settings, respectively. 
Besides Precision, Recall, and F1 score, we also report the average number of sampling iterations (\# Samples) as an efficiency indicator for CoPT.}
\label{tab:cot_acc}
\end{table*}

\begin{highlight}
\textit{\textbf{Finding 3:} Due to the limitations of current reasoning evaluating benchmark, many models are overestimated. Relying solely on answer matching to assess LLMs' reasoning abilities lacks rigor and reliability.}
\end{highlight}
Ensuring the correctness of the reasoning is crucial, but using AI tools like GPT-4 for verification is too expensive. There is a need for a more lightweight method to examine the correctness of the CoT.

\subsubsection{Case Study}

We assume that the model's confidence may change when it makes incorrect assertions. To verify this assumption, we begin with a case study analysis, randomly sampling instances where the model made correct predictions. For each case, we plot a line graph to show changes in the model's confidence over time. 
By comparing each line segment with the corresponding reasoning steps, we observe that significant drops in confidence often occur when the reasoning steps are not supportive of or even contradict the final chosen answer.

Two real cases are shown in Figure~\ref{fig:case_analysis}. In Case 1, the model's confidence drops sharply after the second reasoning step. On the right-hand side, during this step, the model states \textit{the primary cause is the relation between the earth and the moon} (marked in red), aligning with choice (A), which contradicts its final prediction (C). In fact, this statement is indeed incorrect. A similar pattern is also observed in Case 2. Please refer to Appendix~\ref{sec:case_study_sup_app} for additional case studies and numerical analysis.

\subsubsection{CoP with Decision Tree}
To objectively describe this pattern, we train a decision tree to automatically learn classification conditions based on features extracted from the CoP. Specifically, we sample 200 responses from LLaMA-2 7b in the ARC-Challenge dataset and consider GPT-4's annotation as the ground truth.
Then, we extract three features from the model's confidence based on CoP: the maximum value, i.e., $\max(p_1^{v^*}, \cdots, p_k^{v^*})$, the minimum value, i.e., $\min(p_1^{v^*}, \cdots, p_k^{v^*})$, and the minimum change in confidence, i.e., $\min(p_1^{v^*} - p_{0}^{v^*},\cdots,p_k^{v^*} - p_{k-1}^{v^*})$, which could be a large negative number. Using these features and labels, we train a decision tree, namely \textbf{CoP Tree}, with 16 leaf nodes to identify potentially incorrect CoT. 

After training, we use CoP Tree for CoT screening experiments. Specifically, the model repeatedly generates responses to the same question until the decision tree confirms the CoT as correct or the maximum number of samples is reached. We collect the selected response, along with the response obtained under greedy search (GS) settings and 3-time sampling with the CoP score as a control. 
In addition, we compared strong baseline methods Least-to-Most prompting~\cite{zhou2022least} and Tree-of-Thought (ToT)~\cite{yao2023treethoughtsdeliberateproblem}.
We use the annotations from GPT-4 as the ground truth to determine if the CoT contains any errors. Then, we assess the \textbf{CoT Accuracy}, which measures the proportion of responses with completely correct CoT across all questions within the selected responses under these three settings. 

The results are presented in Table~\ref{tab:cot_acc}.
The experiment reveals a 13\% average enhancement in CoT accuracy following CoP Tree filtering of responses. Remarkably, CoP Tree was trained based on confidence features extracted from 200 CoTs generated by the LLaMA-2 7b model, yet it demonstrates outstanding generalization in cross-model predictions. On the other hand, this also suggests a commonality among confidence features in the model. Furthermore, we combine all samples from each model and use GPT-4 annotations as the ground truth to calculate the precision of the decision tree. The results reveal an average precision of 88\%, indicating its high performance in identifying correct CoTs according to the CoP features. In Appendix~\ref{sec:cop_tree_sup_ana} and~\ref{sec:sig_ana_copt_app}, we provide more analysis.

\subsubsection{CoP Tree Error Analysis}

In the classification of the reasoning results of LLaMA-2 7b, the CoP Tree generated 109 false positive samples (incorrect reasoning classified as correct). We conducted a statistical analysis and plotted the corresponding confidence changes of these samples in Appendix~\ref{sec:error_ana_app} Figure~\ref{fig:case_ana_109}. Overall, in the false positive cases, the model exhibited high confidence during reasoning. 89.9\% of false positive cases were early answering cases (marked in orange).  This implies that the model seemed unaware of its errors. This resulted in the confidence change curve being misleading. Please refer to Appendix~\ref{sec:error_ana_app} for more error analysis.

\section{Conclusion}

In this work, we introduce CoP, a probing method to detect changes in confidence during the reasoning process of LLMs. Using CoP, we analyze the necessity and accuracy of CoT. We find that the early answering issue in the LLMs is often due to task simplicity, indicating that CoT is unnecessary for simple tasks. We then propose the CoP Score to identify CoTs that lead to positive improvements. Finally, through case analysis and pattern recognition, we develop the CoP Tree to detect errors in the LLM's reasoning process.

\section*{Limitations}

We summarize the limitations in three points. Firstly, CoP currently only applies to multiple-choice questions or questions where the answer is a single token. This is because the confidence detected by CoP comes from a single token. If the target token set includes words that span more than one token, such as \textit{carbonated}, which is tokenized into [\textit{\_carbon}, \textit{ated}] or even a number like \textit{100}, which is tokenized into [\textit{\_}, \textit{1}, \textit{0}, \textit{0}], defining the model's confidence in the final prediction becomes challenging when the target word exceeds one token. We are exploring the use of the perplexity of multiple tokens instead of the probability of a single token to make CoP applicable to a wider range of tasks. We look forward to presenting more findings in future work.

Secondly, regarding the necessity of CoT, we can only provide a general conclusion: simple tasks do not require CoT. However, it is difficult to determine in advance whether a task is simple, making it impossible to pre-judge whether CoT is needed for a particular question.

Lastly, concerning the accuracy of CoT, the CoP Tree has high precision but relatively low recall. This means it is a strict classifier. Consequently, the resampling strategy of the CoP Tree may reject some correct CoTs, leading to an increase in the number of samples needed.

\section*{Ethic Statement}

In this research, GPT-4 was employed as an evaluator in a manner consistent with ethical guidelines. Transparency about its usage, accountability for its outputs, and mitigation of potential biases were prioritized. Data privacy and security were strictly maintained, and the AI's limitations were acknowledged, ensuring it supplemented rather than replaced human judgment. This approach aimed to enhance the research quality while upholding academic integrity and ethical standards. Refer to Appendix~\ref{sec:gpt4-eval} for the detailed implementation of the GPT-4 evaluation.

\section*{Acknowledgements}
This work was partially supported by Hong Kong RGC GRF No. 14206324, CUHK direct grant No. 4055209, and CUHK Knowledge Transfer Project Fund No. KPF23GWP20.

\bibliography{custom}

\clearpage
\appendix

\section{Experiment Details}
\label{sec:appendix}
\subsection{Detailed Experiment Settings}
\label{sec:settings_app}
We mainly conduct our experiments on the following three datasets:\vspace{-0.2cm}
\begin{itemize}
    \item \textbf{Massive Multitask Language Understanding (MMLU)} includes approximately 16,000 multiple-choice questions across 57 academic subjects, which are categorized into four super categories: STEM, Humanities, Social Science, and Other. ~\cite{hendrycks2021measuring}.\vspace{-0.2cm}
    \item \textbf{AI2 Reasoning Challenge (ARC)} is a set of grade-school science questions, consisting of two subsets: ARC-Easy with 2,376 samples and ARC-Challenge with 1,172 samples. ~\cite{clark2018think}.\vspace{-0.2cm}
    \item \textbf{BIG-Bench Hard Multiple-Choice (BBH MC)} is the multiple-choice subset of BBH~\cite{suzgun2022challenging}, including 4074 samples of 17 tasks from BBH.\vspace{-0.2cm}
\end{itemize}
All experiments use a 5-shot setup. CoT for demonstrations in MMLU and BBH experiments come from the \textit{Chain-of-Thought Hub}~\cite{fu2023chainofthought}, while for ARC experiments, it is manually annotated by our team. For detailed information, including prompting templates and ARC demonstrations, please refer to Appendix~\ref{sec:appendix_prompting_template}. In all experiments, we use half-precision (\texttt{float16}) for inference.
For the experiments using sampling decoding, we set the temperature to 0.7, top-k to 40, and top-p to 0.9.

\subsection{Prompting Template}
\label{sec:appendix_prompting_template}
For all reasoning experiments, we use the following prompting template.
\begin{table}[H]
    \small
    \begin{tcolorbox}
    \textcolor{blue}{\textbraceleft INSTRUCTION\textbraceright}\\
    Question: \textcolor{blue}{\textbraceleft DEMO QUES. 1\textbraceright}\\
    Answer: Let's think step by step. \textcolor{blue}{\textbraceleft DEMO ANS. 1\textbraceright}\\
    \\
    ...\\
    \\
    Question: \textcolor{blue}{\textbraceleft DEMO QUES. 5\textbraceright}\\
    Answer: Let's think step by step. \textcolor{blue}{\textbraceleft DEMO ANS. 5\textbraceright}\\

    Question: \textcolor{blue}{\textbraceleft QUES.\textbraceright}\\
    Answer:
    
    \end{tcolorbox}
    \caption{The prompting template used in experiments. The placeholders within the blue braces need to be filled with corresponding data during the experiment.}
    \label{tab:instructiontemplate1}
\end{table}

For experiments with BBH and MMLU, the instructions and demonstrations used are identical to those used in the Chain-of-Thought hub experiment setup~\cite{fu2023chainofthought}. For example, the instruction for the \textit{tracking shuffled three objects} task is:
\begin{quote}
    \textit{A task requiring determining the final positions of a set of objects given their initial positions and a description of a sequence of swaps.}    
\end{quote}
The instructions and demonstrations used for the ARC dataset are shown in Table~\ref{tab:arc_prompts}.

\subsection{LLMs Checkpoints}
\label{sec:checkpoints}
We provide the detailed version of checkpoints in Table \ref{tab:ckpt}.

\begin{table}[H]
\centering
\begin{tabular}{ll}
\hline
\multicolumn{1}{c}{\textbf{LLM}} & \multicolumn{1}{c}{\textbf{Checkpoints}} \\ \hline
Vicuna 7b                           & \texttt{vicuna-7b-v1.5}                         \\
Mistral 7b                           & \texttt{Mistral-7B-Instruct-v0.3}                         \\
LLaMA-3 8b                           & \texttt{Meta-Llama-3-8B-Instruct}                         \\
Qwen2 7b                           & \texttt{Qwen2-7B-Instruct}                         \\
LLaMA-2 7b                           & \texttt{llama-2-7b-chat}                        \\
LLaMA-2 13b                           & \texttt{llama-2-13b-chat}                        \\
LLaMA-2 70b                           & \texttt{llama-2-70b-chat}                        \\
GPT-4                            & \texttt{gpt-4-turbo-2024-04-09}                             \\ \hline

\end{tabular}
\caption{LLMs involved in the experiments and the corresponding checkpoints.}
\label{tab:ckpt}
\end{table}

\subsection{GPT-4 Evaluation}
\label{sec:gpt4-eval}
\subsubsection{Instructions for GPT-4}
We use GPT-4 as an evaluator to examine whether the CoT generated by the model contains potential errors (refer to Table~\ref{tab:ckpt} for the detailed version). Briefly, we ask GPT-4 to check for contradictory statements or statements that do not align with facts in the CoT. For the instructions, refer to Table~\ref{tab:gpt4-instruction}.

\subsubsection{Human Evaluation}
To ensure the reliability of GPT-4's evaluation, we randomly selected 50 samples from the responses of all models on ARC-Challenge for manual annotation. We recruited four PhD students to annotate the golden answers of whether both CoT and final answers are correct. All the answers are consistently validated by all four annotators.
The Pearson correlation coefficient between GPT-4's annotations and the golden answers is \textbf{0.879}, proving that GPT-4 is a capable evaluator.

\begin{table}[t]
    \small
    \begin{tcolorbox}
    \#\#\# Instruction\\ \\
    Below is a question paired with an answer provided by an AI assistant. With reference to the ground truth, you are tasked with examining the assistant's response. The primary focus of your evaluation should be to discern whether there are any inconsistencies or assertions within the reasoning that contradict established facts, rather than merely assessing the correctness of the answer itself. If no issues are found in the response, reply with \\ \\
    <Evaluation Results: Pass> \\ \\
    Should you identify problematic statements, please indicate the portions of its answer that are problematic and respond in the following format: \\ \\
    <Evaluation Results: Fail> \\ \\
    <Start of problematic description 1> \\
    ... \\
    <End of problematic description 1> \\  \\
    <Start of problematic description 2> \\
    ... \\
    <End of problematic description 2> \\
    ... \\ \\
    \#\#\# Input\\ \\
    Question: \textcolor{blue}{\textbraceleft QUES. \textbraceright} \\
    AI Answer: \textcolor{blue}{\textbraceleft ANS. \textbraceright} \\ \\
    \#\#\# Output    
    \end{tcolorbox}
    \caption{The prompting template for GPT-4 evaluation.}
    \label{tab:gpt4-instruction}
\end{table}

\begin{table*}[]
    \centering
    \includegraphics[trim={0cm 0cm 0cm 0cm}, clip, width=1.0\textwidth]{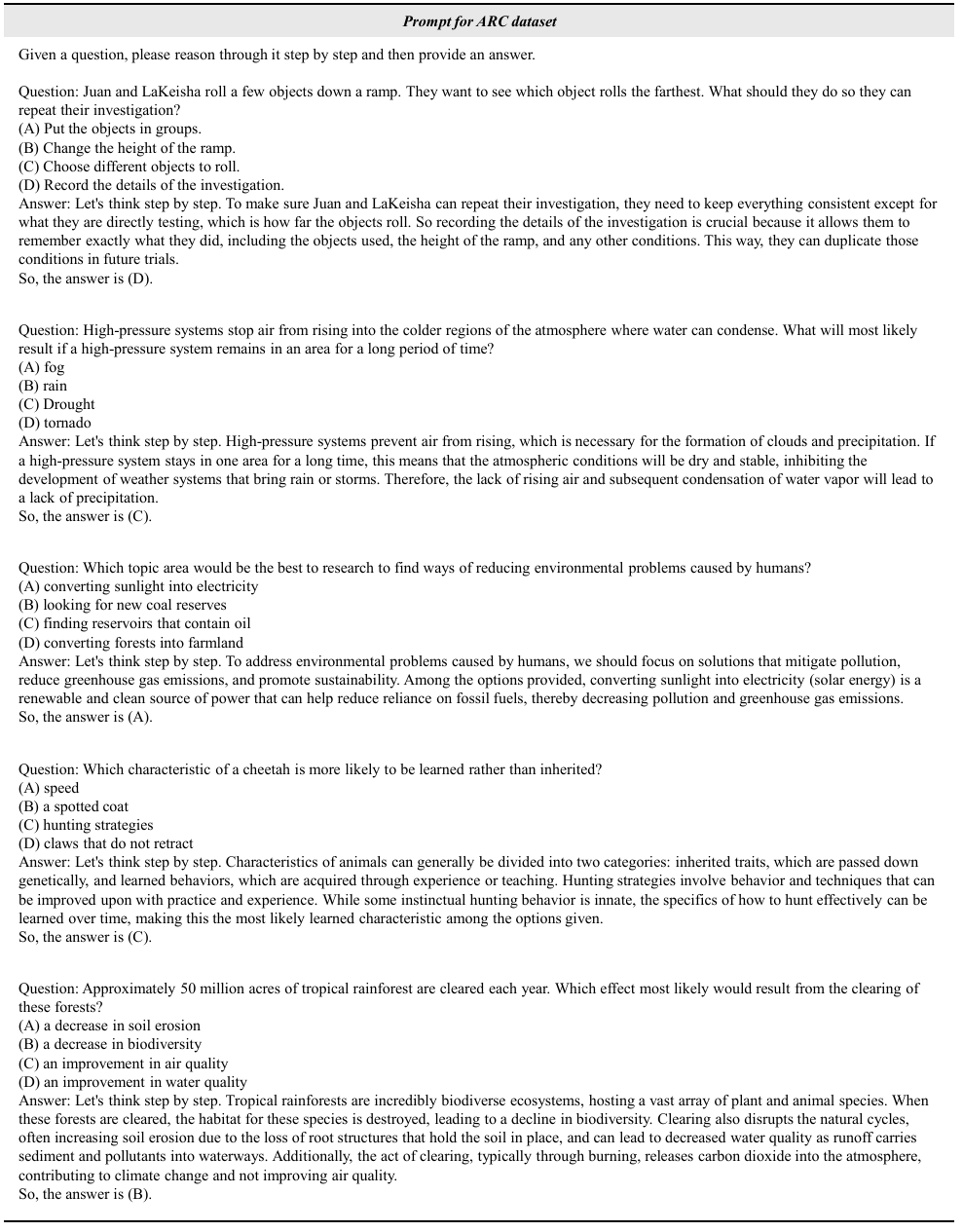}
    \caption{The prompts used for reasoning on ARC dataset.}
    \label{tab:arc_prompts}
\end{table*}

\section{Experiments Results Supplement}
\subsection{Impact of Problem Difficulty on EAR Performance}
\label{sec:app_arc_ear}
In addition to using model accuracy as a criterion for task difficulty, we further investigated the variation of EAR under human-annotated difficulty levels. The ARC dataset consists of two subsets: Easy and Challenge, with each subset containing problems spanning from grade 3 to grade 9. The EAR results across subsets are presented in Table~\ref{tab:arc_ear}.

\begin{table}[h]
\centering
\begin{tabular}{ccc}
\hline
\multicolumn{1}{l}{\textbf{Grade}} & \multicolumn{1}{l}{\textbf{ARC-Challenge}} & \multicolumn{1}{l}{\textbf{ARC-Easy}} \\ \hline
3                                  & 0.7399                                     & 0.8323                                \\
4                                  & 0.7497                                     & 0.8642                                \\
5                                  & 0.7502                                     & 0.8578                                \\
6                                  & 0.6901                                     & 0.8263                                \\
7                                  & 0.7023                                     & 0.8017                                \\
8                                  & 0.6665                                     & 0.8024                                \\
9                                  & 0.6684                                     & 0.8153                                \\ \hline
\end{tabular}
\caption{Comparison of EAR across different difficulty levels in the ARC dataset.}
\label{tab:arc_ear}
\end{table}

The analysis reveals that the ARC-Challenge subset indeed shows a lower EAR, which provides additional support for the findings reported in the main text. Interestingly, we observed a general downward trend in EAR from grade 3 to grade 9. While this trend is not strictly monotonic, it aligns with our expectations.

\subsection{Correlation Analysis}
\label{sec:corr_ana_app}
From Figure~\ref{fig:ear_acc} and Figure~\ref{fig:cot_score_acc}, we can see that there is a positive correlation between EAR and accuracy, as well as between CoPS and accuracy. To validate this observation, we further conducted numerical analysis. We calculated the Pearson correlation coefficients between EAR and accuracy, and between CoPS and accuracy, respectively. The results are reported in Table~\ref{tab:acc_corr}. 

\begin{table}[h]
\centering\small
\setlength\tabcolsep{4pt}
\begin{tabular}{lcccc}
\hline
                     & \multicolumn{2}{c}{\textbf{EAR \& Acc}} & \multicolumn{2}{c}{\textbf{CoPS \& Acc}} \\
\textbf{Model}       & \textbf{MMLU}       & \textbf{BBH}      & \textbf{MMLU}       & \textbf{BBH}       \\ \hline
\textbf{Vicuna 7b}   & 0.9041              & 0.8303            & 0.9293              & 0.9923             \\
\textbf{LLaMA-2 7b}  & 0.8814              & 0.8240            & 0.8455              & 0.9620             \\
\textbf{LLaMA-2 13b} & 0.9011              & 0.7233            & 0.7892              & 0.9306             \\
\textbf{LLaMA-2 70b} & 0.8794              & 0.6874            & 0.8939              & 0.9876             \\
\rowcolor[HTML]{EFEFEF}\textbf{Avg.}        & 0.8915              & 0.7663            & 0.8645              & 0.9681             \\ \hline
\end{tabular}
\caption{Correlation analysis between EAR and accuracy, and CoPS and accuracy.}
\label{tab:acc_corr}
\end{table}

The correlation between EAR and Accuracy is strong, with average Pearson correlation coefficients of 0.8915 and 0.7663 on the MMLU and BBH datasets, respectively. On the other hand, CoPS and Accuracy also exhibit strong positive correlation, averaging 0.8645 on MMLU and 0.9681 on BBH.

\begin{figure*}[t]
    \centering
    \includegraphics[width=\linewidth]{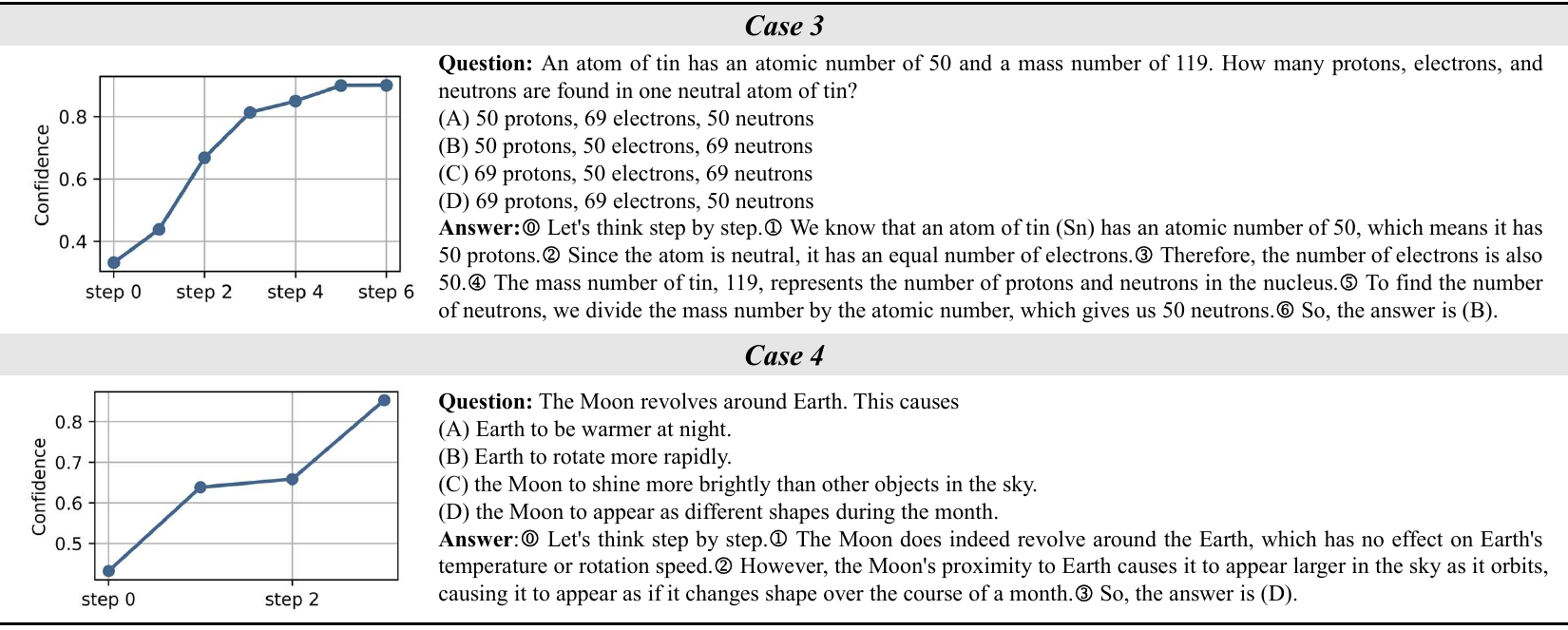}
    \caption{Analysis of two real cases: The line plot shows changes in the model's confidence in its final prediction over the reasoning steps. On the right are the corresponding cases, including the question and the response generated by the model. We add numerical indices to indicate probing points, helping readers match the line plot with the CoT.}
    \label{fig:case_analysis_2}
\vspace{-0.3cm}
\end{figure*}

\subsection{Positive \& Negative Effect of CoT}
\label{sec:pos_eng_cot}

We analyzed the proportion of answers that changed from false to true and from true to false after the model's reasoning on the MMLU and BBH datasets. Specifically, in our CoP, the results obtained from probing at step 0 are considered the model's predictions without any reasoning, i.e., $j_0$. The final generated answers are considered the model's predictions after full reasoning, i.e., $j^*$. Hence, we define the effect of CoT as the following equation,
\begin{equation}
    \mathrm{Effect} = 
\begin{cases} 
\mathrm{Pos.} & \text{if } j_0\text{ is False and } j^*\text{ is True}\\
\mathrm{Neg.} & \text{if } j_0\text{ is True and } j^*\text{ is False}
\end{cases}
\label{eq:neg_pos_cot}
\end{equation}

We further compute the ratio of positive CoT and negative CoT and present the results in Figure~\ref{fig:T2FandF2T}.
Overall, the positive effects of CoT are more evident, although there are still a number of cases where the model's reasoning changes a correct answer to an incorrect one.

\subsection{Case Study Supplement}
\label{sec:case_study_sup_app}
Figure~\ref{fig:case_analysis_2} shows two cases of reasoning where confidence increases. In case 3, as the model's confidence in its reasoning grows, it seems more certain about its choices, but the reasoning process is incorrect. The number of neutrons is found by subtracting the atomic number from the mass number, not by division. In this case, the change in the model's confidence aligns with what we would expect from a "good" thinking pattern, yet it leads to an erroneous reasoning. We attribute this error to the limited reasoning ability and knowledge of the model. It is not aware that its reasoning is incorrect. In case 4, confidence also increases as the reasoning progresses, and this case is an example where there are no errors and the reasoning is completely correct.

We believe that it is effective to judge whether errors occur in the model's reasoning process based on the pattern of confidence changes. We can identify some patterns that roughly represent correct and incorrect reasoning. It is undeniable that this approach is not entirely accurate. When the model is not aware of its mistakes, confidence does not change, leading to misjudgments.

To further validate the effectiveness of the idea of pattern-matching, we conducted additional statistical analysis.
Based on the results of the CoP, we first identify the bad pattern that reasoning steps that lead to a significant decrease in confidence. We then compare the annotations and perform statistical analysis based on whether the pattern matches and whether the step is correct. The results of LLaMA-2 7b are as follows:

\begin{table}[h]
\centering
\small
\begin{tabular}{ccc}
\hline
                                     & \textbf{Match} & \textbf{Mismatch} \\ \hline
\multicolumn{1}{l}{\textbf{Correct}} & 1.54\%         & 72.39\%           \\
\textbf{Incorrect}                   & 10.03\%        & 16.02\%           \\ \hline
\end{tabular}
\caption{The ratio of correct and incorrect reasoning steps that match or do not match the pattern of "decreasing confidence".}
\end{table}

Here is a further explanation of the indicators with an example: Match \& Incorrect - this reasoning step matches the pattern of a significant decrease in confidence and is also incorrect. The values in the table represent the proportion of this kind of sample. Please note that we only conducted statistical analysis on the examples that included incorrect reasoning steps.

These results indicate that the majority of reasoning processes that fit this pattern are incorrect (10.03\% vs. 1.54\%). This suggests that using this pattern to filter out incorrect reasoning steps is reliable. On the other hand, there are still many incorrect reasoning steps that do not fit this pattern (16.02\%). This indicates that this pattern is just one of the patterns for incorrect reasoning and cannot represent the overall trend of confidence changes in all incorrect reasoning. This is consistent with the conclusion drawn from our case study, which is why we need to train a decision tree to further comprehensively describe the patterns of errors.

\subsection{Supplement Analysis of CoP Tree}
\label{sec:cop_tree_sup_ana}
In Table~\ref{tab:cot_acc}, when comparing CoPT and CoPS@3, we found that Vicuna 7b can achieve an absolute improvement of over 12\%. Compared to the 5\%-8\% improvement in other models, the improvement of Vicuna 7b can be considered significant. This may be because Vicuna 7b itself has a lower accuracy (only 55.3\%, while other models are above 60\% and even as high as 76.8\%). 

\begin{figure}[t]
    \centering
    \includegraphics[width=\linewidth]{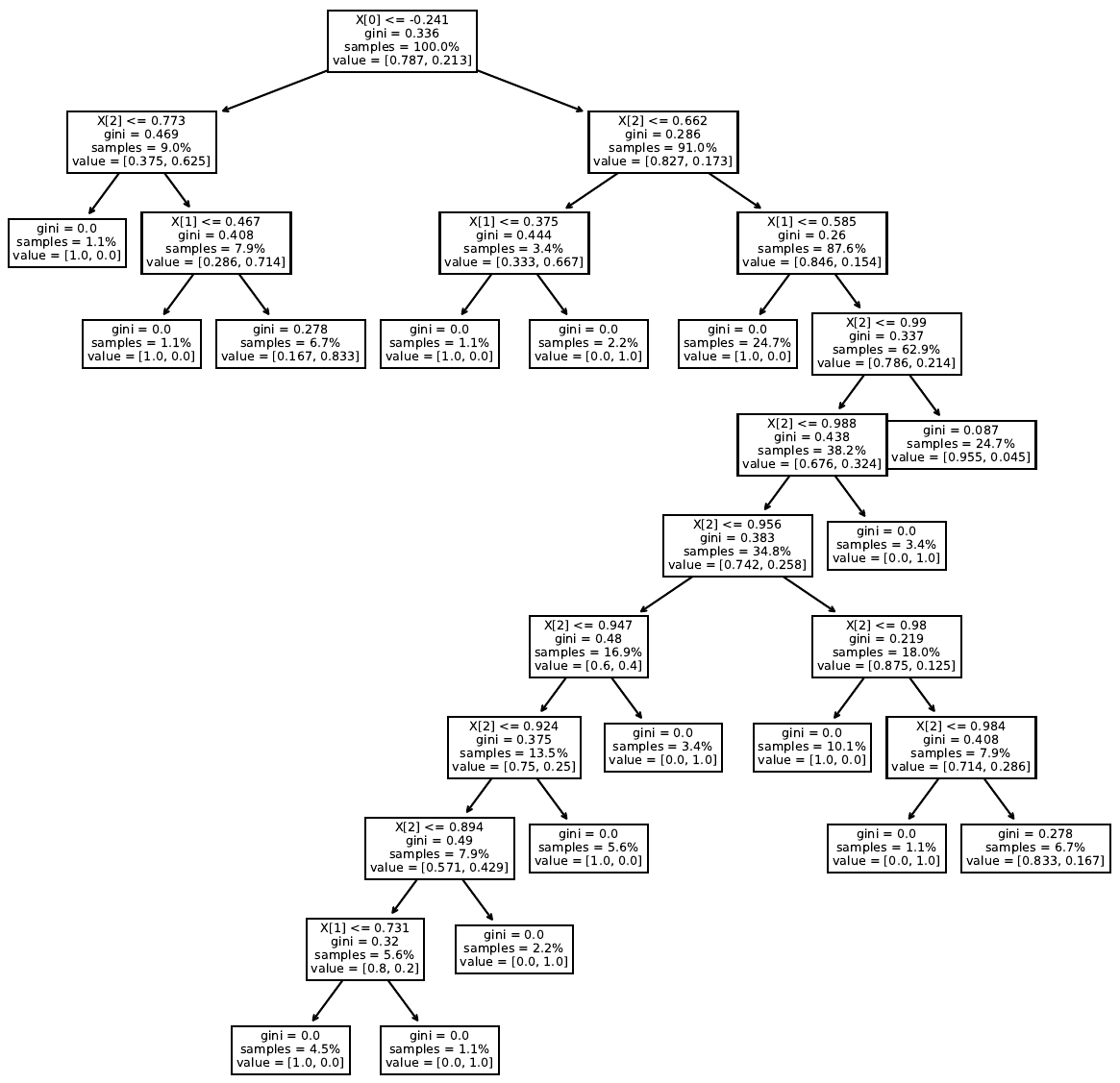}
    \caption{The classification process of the trained CoP Tree, where x[0] is the minimum change in confidence, x[1] is the minimum value, x[2] is the maximum value.
}
    \label{fig:dst}
\end{figure}

Figure ~\ref{fig:dst} illustrates the classification process of the trained CoP Tree. Overall, the decision tree is asymmetric and has a skewed shape. The right subtree of the root node has been split multiple times based on feature 2, which represents the maximum value of confidence. This suggests that the feature is more valuable for distinguishing, but it also raises the possibility that the decision tree may have overfit on this feature.

We further investigated the improvement in answer accuracy brought by CoPT and reported the results in Table~\ref{tab:copt_ans_acc}. The results show that CoPT improved the model's final prediction accuracy. In comparison, CoPT demonstrated a more clear improvement in CoT accuracy.

\begin{table*}[t]
\centering
\begin{tabular}{lcccc}
\hline
\textbf{Model}       & \textbf{GS Answer Acc.} & \textbf{CoPT Answer Acc} & \textbf{GS CoT Acc.} & \textbf{CoPT CoT Acc} \\ \hline
\textbf{Vicuna 7b}   & 66.9                    & 74.3                     & 49.1                 & 67.6                  \\
\textbf{Mistral 7b}  & 82.2                    & 86.7                     & 63.7                 & 76.4                  \\
\textbf{LLaMA-3 8b}  & 86.1                    & 88.3                     & 67.4                 & 78.9                  \\
\textbf{Qwen2 7b}    & 88.7                    & 91.6                     & 71.4                 & 82.8                  \\
\textbf{LLaMA-2 7b}  & 69.2                    & 74.6                     & 55.4                 & 68.7                  \\
\textbf{LLaMA-2 13b} & 73.2                    & 79.4                     & 60.5                 & 72.8                  \\
\textbf{LLaMA-2 70b} & 84.0                    & 90.0                     & 66.9                 & 81.1                  \\
\rowcolor[HTML]{EFEFEF}\textbf{Avg.}        & 78.6                    & 83.6                     & 62.1                 & 75.5                  \\ \hline
\end{tabular}
\caption{Answer accuracy improvement achieved by CoP Tree. For ease of comparison, we also included the CoT Accuracy from the main text.}
\label{tab:copt_ans_acc}
\end{table*}

\subsection{Significance Analysis of CoP Tree}
\label{sec:sig_ana_copt_app}
 Statistical evaluation of the CoPT method against the ToT baseline yielded compelling results. The paired t-test analysis revealed that CoPT (Mean=75.47\%) consistently outperformed the ToT (Mean=70.17\%), with a mean performance gain of 5.3\%. The observed difference achieved statistical significance (p < 0.05), accompanied by a robust effect size (Cohen's d = 0.82). This quantitative evidence substantiates CoPT's superior performance in the experimental setting.

Moreover, we employed Bayes Factor analysis to evaluate classifier performance. The null hypothesis was set as the performance of a random classifier, which theoretically achieves 50\% precision in binary classification tasks. This baseline represents purely random guessing and was set as the theoretical value for all seven test models. The alternative hypothesis was based on our classifier's actual performance across seven independent models. The Bayes Factor analysis yielded a $\text{BF}_{10}$ of approximately 4.054e+04, indicating overwhelming evidence in support of the alternative hypothesis, effectively ruling out the possibility of random classification performance.
\subsection{Error Analysis of CoP Tree}
\label{sec:error_ana_app}
\begin{figure*}[t]
    \centering
    \includegraphics[width=\linewidth]{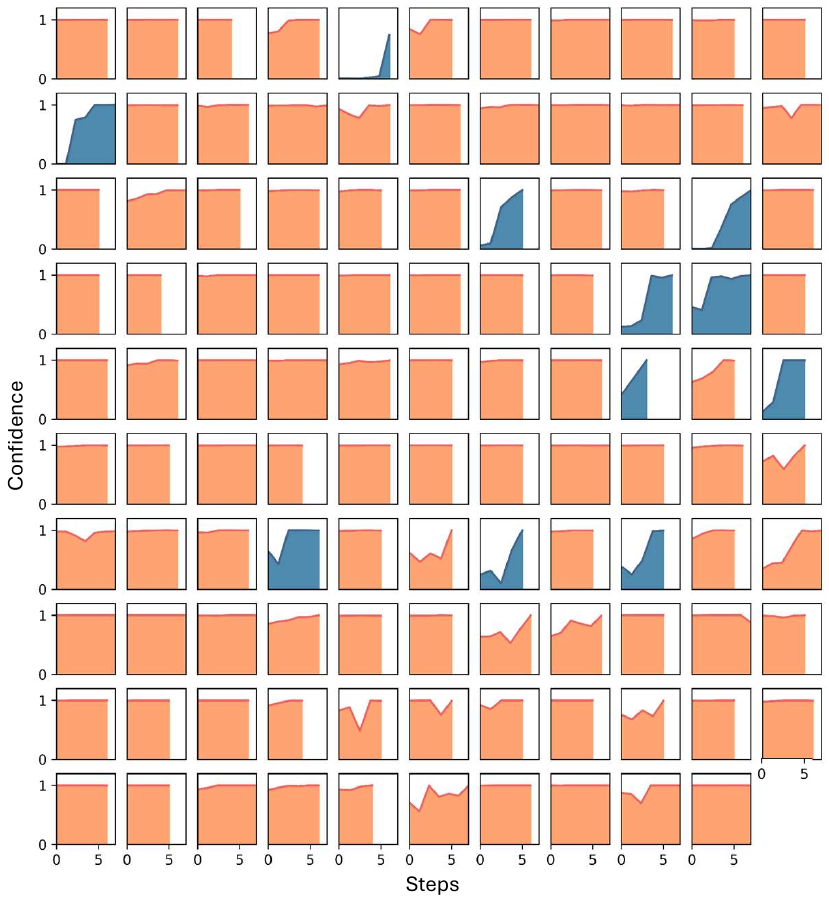}
    \caption{The confidence change corresponding to the 109 false positive cases misclassified by the CoP Tree. The orange color represents early answering cases, while the blue color represents cases that are not early answering.}
    \label{fig:case_ana_109}
\end{figure*}

On the ARC-Challenge dataset, we collected the CoT and CoP generated by LLaMA-2 7b under the setting of greedy search. We used the CoP Tree to classify these CoPs and compared them with the labels generated by GPT-4. Finally, we collected 109 False Positive cases, where the CoP Tree believed there was no error, but there actually was. We plotted the confidence changes of these 109 cases into a line graph (see Figure~\ref{fig:case_ana_109}).

Overall, in the false positive cases, the model exhibited high confidence during reasoning. We further divided these cases based on whether they belonged to the phenomenon of early answering and found that 89.9\% of false positive cases were early answering cases (marked in orange). This implies that the model seemed unaware of its errors. We conducted manual checks on false positive cases that did not belong to early answering. In these cases, we found that the model often made errors at the beginning of reasoning (resulting in low confidence). The model seemed to convince itself to accept the error during the subsequent reasoning process, leading to increasing confidence.

From the perspective of confidence changes, these two types of errors align with our definition of "good" thinking patterns. For example, maintaining high confidence throughout the reasoning process, or increasing confidence as the reasoning progresses. When the model is unable to realize its mistakes, or convinces itself to accept errors during reasoning, the changes in confidence can be quite misleading. This highlights the limitations of CoP Tree, which relies on model confidence to determine the correctness of reasoning.

\subsection{Tree-of-Thought Experiment Setup and Results Supplement}
\label{sec:tot_app}
We referred to the original paper and source code of ToT\footnote{https://github.com/princeton-nlp/tree-of-thought-llm}, and designed a BFS algorithm for the ARC-Challenge dataset. Initially, we counted the reasoning steps of the standard answers in the dataset. Most of the reasoning can be completed within 6 steps. Therefore, we defined the maximum depth of ToT as 6. We specified that the model samples 5 responses for each reasoning. Then, through self-evaluation, the model selects the optimal reasoning process from the five responses (ToT width is 1). Repeat this reasoning process until the model gets the answer or reaches the maximum tree depth. The prompts used in the ToT experiment are shown in Table~\ref{tab:tot_propose_prompts} and Table~\ref{tab:tot_vote_prompts}.

The experimental results are shown in Table~\ref{tab:cot_acc}. The improvement in reasoning ability of LLM by ToT varies significantly: Mistral 7b reasoning with ToT shows a 14.9\% improvement compared to greedy search, while Vicuna 7b's accuracy decreases by 0.4\%. We conducted extensive case studies and found that the improvement brought by ToT highly depends on the model's ability to follow instructions. ToT breaks down tasks, requiring the model to complete specific subtasks step by step, such as reasoning the next step or evaluating and selecting the optimal reasoning step. However, these subtasks often dictate the model's output format. If the model's output format does not fully comply with the instructions, the predefined regular matching program may fail to effectively extract key information, leading to breaks in the ToT reasoning process and affecting the generation of the final answer. For example, when the model self-evaluates and selects candidate responses, we specify that it should summarize its final choice as \textit{"The best choice is candidate \{idx\}."} If the model does not follow this instruction, for instance, if it responds with \textit{"The best choice is the first one,"} the subsequent matching program will not be able to extract the evaluation results.

Adding more human input may alleviate the issue, such as testing better prompts, optimizing regular expressions to make them more general, etc., but it cannot fundamentally solve the problem. The performance decline caused by the model deviating from instructions sometimes exceeds the improvement in reasoning ability brought by ToT. This is why, for weak models, the improvement brought by ToT is not significant and may even have negative effects.

On the other hand, ToT brings significant improvements to strong models, such as Mistral 7b and LLaMA-3 8b, surpassing the improvements brought by CoPT. We believe the main reason is that The ToT algorithm includes a step-levl error correction process. ToT requires the model to self-evaluate and select better reasoning steps, allowing for timely correction of errors and preventing error accumulation. In contrast, CoPT evaluates the correctness after the model completes all reasoning. Therefore, every time a correction is made, the model needs to reason from beginning again. During the process of re-reasoning, the model may generate new errors. This is a limitation of the sample-level error correction mechanism. The performance improvement brought by ToT comes at the cost of significant inference overhead. Compared to CoPT, under the 5-shot setting, ToT consumes 6-7 times more tokens than CoPT. We counted the total number of tokens consumed during model inference under both settings, which includes the sum of input tokens and output tokens. We reported the results in Table~\ref{tab:token_count}.

\begin{table}[t]
\centering
\small
\begin{tabular}{lll}
\hline
\multicolumn{1}{c}{\textbf{Model}} & \multicolumn{1}{c}{\textbf{CoPT}} & \multicolumn{1}{c}{\textbf{ToT}} \\ \hline
Vicuna 7b                          & 5,931,135                         & 36,262,747                       \\
Mistral 7b                         & 5,400,881                         & 34,532,024                       \\
LLaMA-3 8b                         & 5,054,359                         & 31,855,485                       \\
Qwen2 7b                           & 4,539,389                         & 28,143,109                       \\
LLaMA-2 7b                         & 6,079,912                         & 39,035,407                       \\
LLaMA-2 13b                        & 5,390,422                         & 33,522,586                       \\
LLaMA-2 70b                        & 5,641,564                         & 34,282,840                       \\
\rowcolor[HTML]{EFEFEF}Avg.                               & 5,433,952                         & 33,947,743                       \\ \hline
\end{tabular}
\caption{Comparison of inference token consumption on the ARC-Challenge dataset under the settings of CoPT and ToT.}
\label{tab:token_count}
\end{table}

Therefore, developers need to carefully balance the trade-off between overhead and performance. From a cost perspective, CoPT achieves similar performance to ToT at a much lower cost, making it a better choice.

\begin{table}[]
    \centering
    \includegraphics[trim={0cm 0cm 0cm 0cm}, clip, width=0.48\textwidth]{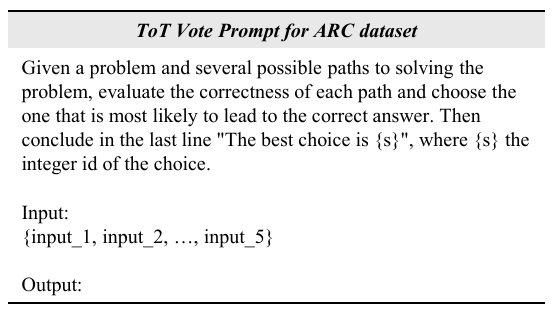}
    \caption{ToT vote prompt used for ARC dataset.}
    \label{tab:tot_vote_prompts}
\end{table}

\begin{table*}[]
    \centering
    \includegraphics[trim={0cm 0cm 0cm 0cm}, clip, width=1.0\textwidth]{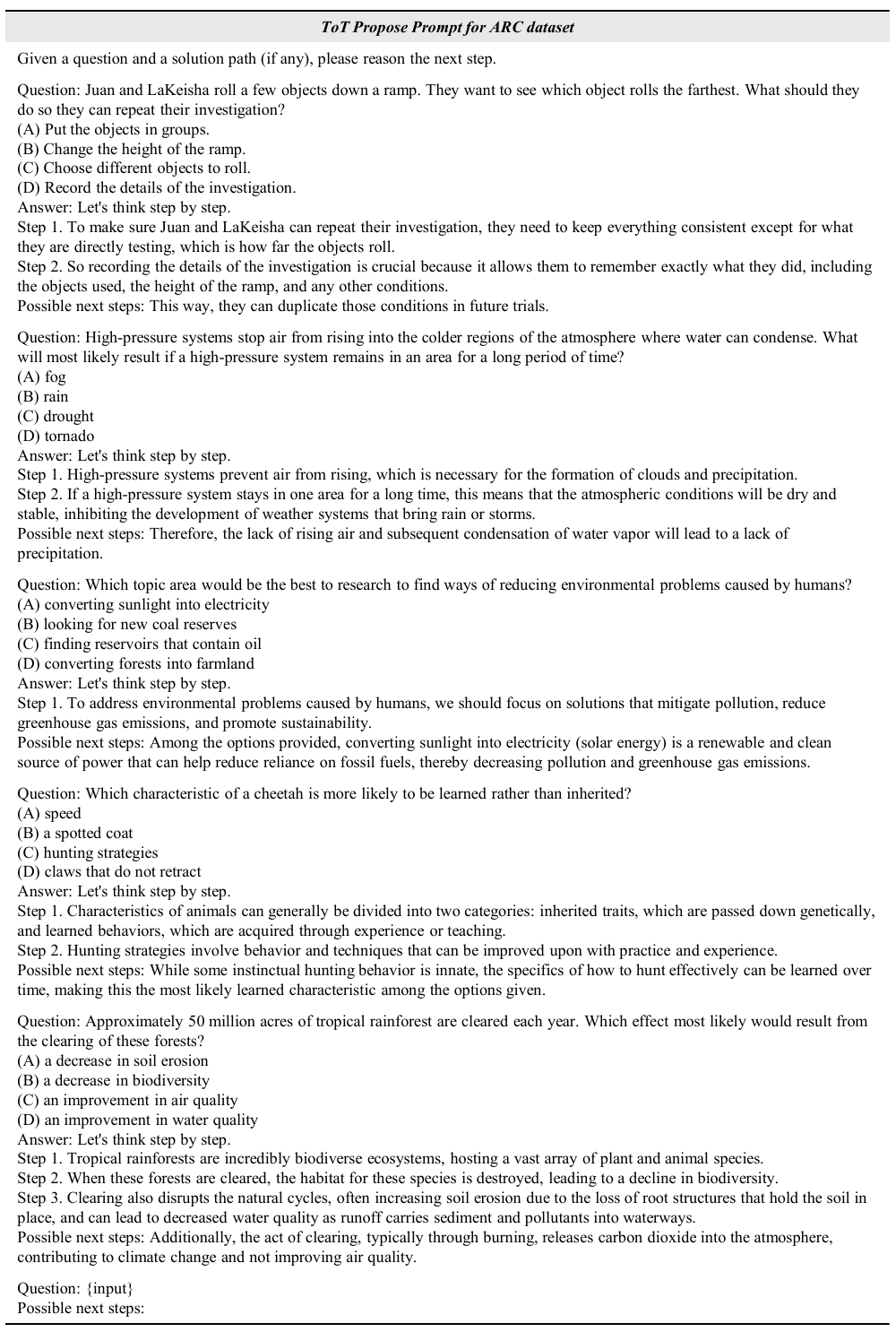}
    \caption{ToT propose prompt used for ARC dataset.}
    \label{tab:tot_propose_prompts}
\end{table*}

\subsection{Least-to-Most Prompting Experiment Setup and Results Supplement}
\label{sec:l2m_app}
We referred to the original paper~\cite{zhou2022least} to create the prompt. We rephrased all five examples used for inference into a task decomposition solving approach, with detailed prompts provided in Table~\ref{tab:least_to_most_prompts}. The experimental results are presented in Table~\ref{tab:cot_acc} within the main text.

Despite the Least-to-Most prompting method not involving reflection and error correction mechanisms, its accuracy in the reasoning process has significantly improved compared to the baseline of CoT + greedy search. This improvement is attributed to the task decomposition mechanism of Least-to-Most Prompting. By breaking down complex problems into sub-problems, the difficulty is reduced, enabling the model to solve simpler sub-problems with fewer errors. This, in turn, enhances the accuracy of the reasoning process. On average, the accuracy of the Least-to-Most Prompting method is comparable to CoPS@3. It is important to note that these two methods are not mutually exclusive. We can generate multiple inference processes by repeatedly sampling using the Least-to-Most Prompting method, calculate the CoP Score, and then select the best inference process.

\begin{table*}[]
    \centering
    \includegraphics[trim={0cm 0cm 0cm 0cm}, clip, width=1.0\textwidth]{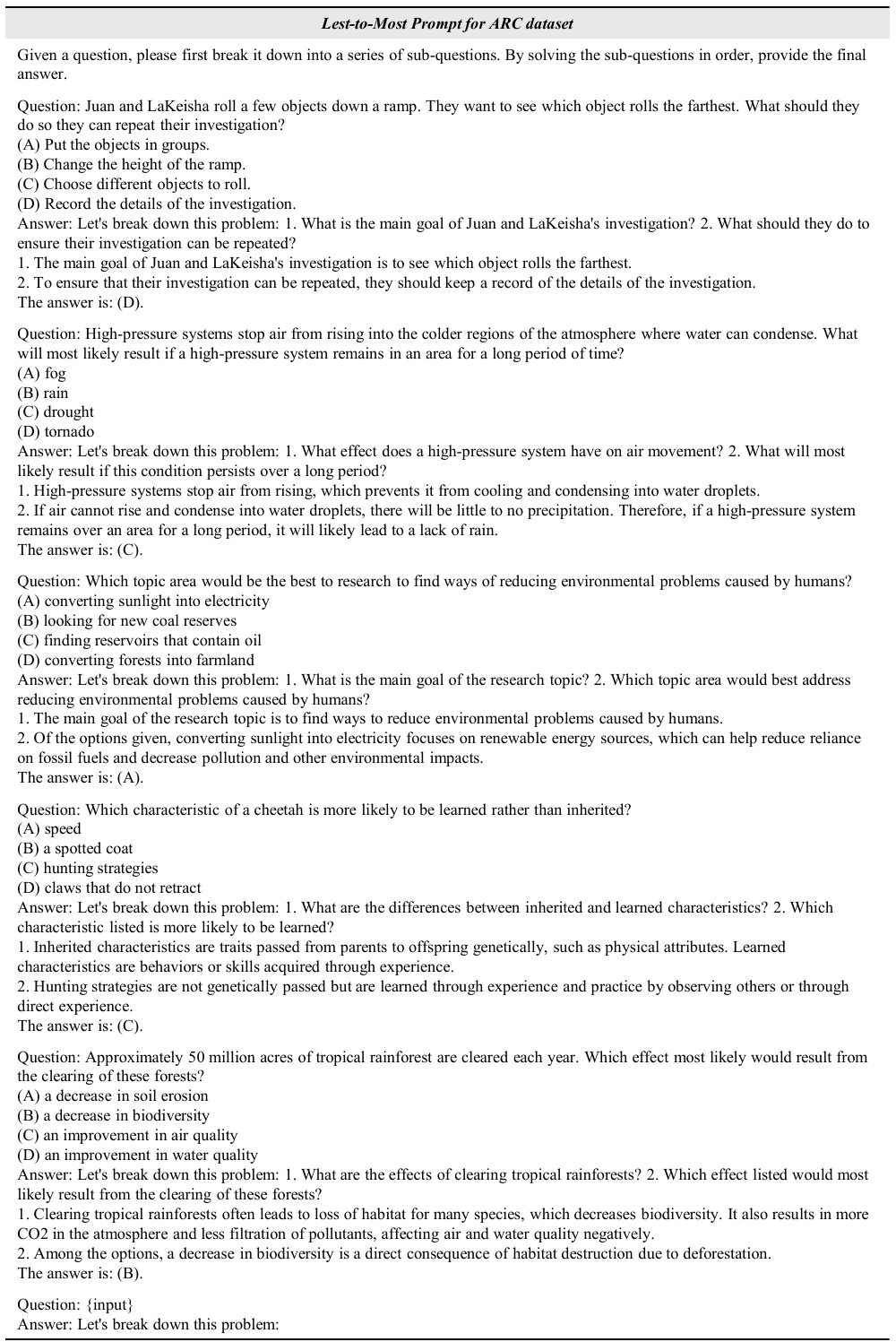}
    \caption{Least-to-Most prompt used for ARC dataset.}
    \label{tab:least_to_most_prompts}
\end{table*}


\end{document}